\documentclass[journal]{IEEEtran}

\usepackage{amsmath}
\usepackage{amssymb}
\usepackage{bm}
\usepackage{color}
\usepackage{environ}
\usepackage{epsfig}
\usepackage{graphicx}
\usepackage{pifont}
\usepackage{times}

\newcommand{\RNum}[1]{\uppercase\expandafter{\romannumeral #1\relax}} 
\newcommand{\Rnum}[1]{\lowercase\expandafter{\romannumeral #1\relax}} 
\newcommand{\Paragraph}[1]{\vspace{-0mm} \noindent \textbf{#1} \hspace{0mm}}

\ifCLASSINFOpdf
\else
\fi

\begin{document}

\title{Multi-Task Convolutional Neural Network for \\ Pose-Invariant Face Recognition}

\author{Xi Yin\IEEEcompsocitemizethanks{ Xi Yin and Xiaoming Liu are with the Department of Computer Science and Engineering, Michigan State University, East Lansing, MI 48824. Corresponding author: Xiaoming Liu, liuxm@cse.msu.edu} and Xiaoming Liu~\IEEEmembership{Member,~IEEE,}}

\maketitle

\begin{abstract}
This paper explores multi-task learning (MTL) for face recognition. 
We answer the questions of how and why MTL can improve the face recognition performance. 
First, we propose a multi-task Convolutional Neural Network (CNN) for face recognition where identity classification is the main task and pose, illumination, and expression estimations are the side tasks. 
Second, we develop a dynamic-weighting scheme to automatically assign the loss weight to each side task, which is a crucial problem in MTL.
Third, we propose a pose-directed multi-task CNN by grouping different poses to learn pose-specific identity features, simultaneously across all poses.
Last but not least, we propose an energy-based weight analysis method to explore how CNN-based MTL works. 
We observe that the side tasks serve as regularizations to disentangle the variations from the learnt identity features.
Extensive experiments on the entire Multi-PIE dataset demonstrate the effectiveness of the proposed approach. 
To the best of our knowledge, this is the first work using all data in Multi-PIE for face recognition. 
Our approach is also applicable to in-the-wild datasets for pose-invariant face recognition and achieves comparable or better performance than state of the art on LFW, CFP, and IJB-A datasets. 
\end{abstract}

\begin{IEEEkeywords}
multi-task learning, pose-invariant face recognition, CNN, disentangled representation
\end{IEEEkeywords}

\IEEEpeerreviewmaketitle

\section{Introduction}
\IEEEPARstart{F}{ace} recognition is a challenging problem that has been studied for decades in computer vision. 
The large variations in Pose, Illumination, Expression (PIE), and etc. will increase the intra-person variation that will challenge any state-of-the-art face recognition algorithms. 
Recent CNN-based approaches mainly focus on exploring the effects of $3D$ model-based face alignment~\cite{taigman2014deepface}, larger datasets~\cite{taigman2014deepface, schroff2015facenet}, or new loss functions~\cite{sun2014deep, schroff2015facenet, littwin2016multiverse, wen2016discriminative} on face recognition performance. 
Most existing methods consider face recognition as a single task of extracting robust identity features. 
We believe that face recognition is not an isolated problem --- often tangled with other tasks. 
For example, when presented with a face image, we will instinctively recognize the identity, pose, and expression at the same time. 
This motivates us to explore multi-task learning for face recognition. 

\begin{table*}
\small
\begin{center}
\caption{\small Comparison of the experimental settings that are commonly used in prior work on Multi-PIE. (* The $20$ images consist of $2$ duplicates of non-flash images and $18$ flash images. In total there are $19$ different illuminations.)}
\label{tab_multipie}
\begin{tabular}{ ccccccccc}
\hline 
setting & session & pose & exp & illum & train subjects / images & gallery / probe images & total & references \\ \hline \hline
\RNum{1} & $4$ & $7$ & $1$ & $1$ & $200$ / $5,383$ & $137$ / $2,600$ & $8,120$ & ~\cite{asthana2011fully, li2012morphable}\\ 
\RNum{2} & $1$ & $7$ & $1$ & $20$ & $100$ / $14,000$ & $149$ / $20,711$ & $34,860$ & ~\cite{zhu2013deep, yim2015rotating} \\ 
\RNum{3} & $1$ & $15$ & $1$ & $20$ & $150$ / $45,000$ & $99$ / $29,601$ & $74,700$ & ~\cite{xiong2015conditional} \\ 
\RNum{4} & $4$ & $9$ & $1$ & $20$ & $200$ / $138,420$ & $137$ / $70,243$ & $208,800$ & ~\cite{zhu2014multi, yim2015rotating} \\ \hline
ours & $4$ & $15$ & $6$ & $20$* & $200$ / $498,900$ & $137$ / $255,163$ & $754,200$ & \\ \hline \hline
\end{tabular}
\end{center}
\vspace{-2mm}
\end{table*}

Multi-task learning (MTL) aims to learn several tasks {\it simultaneously} to boost the performance of the main task or all tasks.
It has been successfully applied to face detection~\cite{chen2014joint,zhang2014improving}, face alignment~\cite{zhang2014facial}, pedestrian detection~\cite{tian2015pedestrian}, attribute estimation~\cite{abdulnabi2015multi}, and so on. 
Despite the success of MTL in various vision problems, there is a lack of comprehensive study of MTL for face recognition.
In this paper, we study face recognition as a multi-task problem where identity classification is the main task with PIE estimations being the side tasks. 
The goal is to leverage the side tasks to improve the performance of face recognition.

\begin{figure}[t]
\begin{center}
\includegraphics[width=0.48\textwidth]{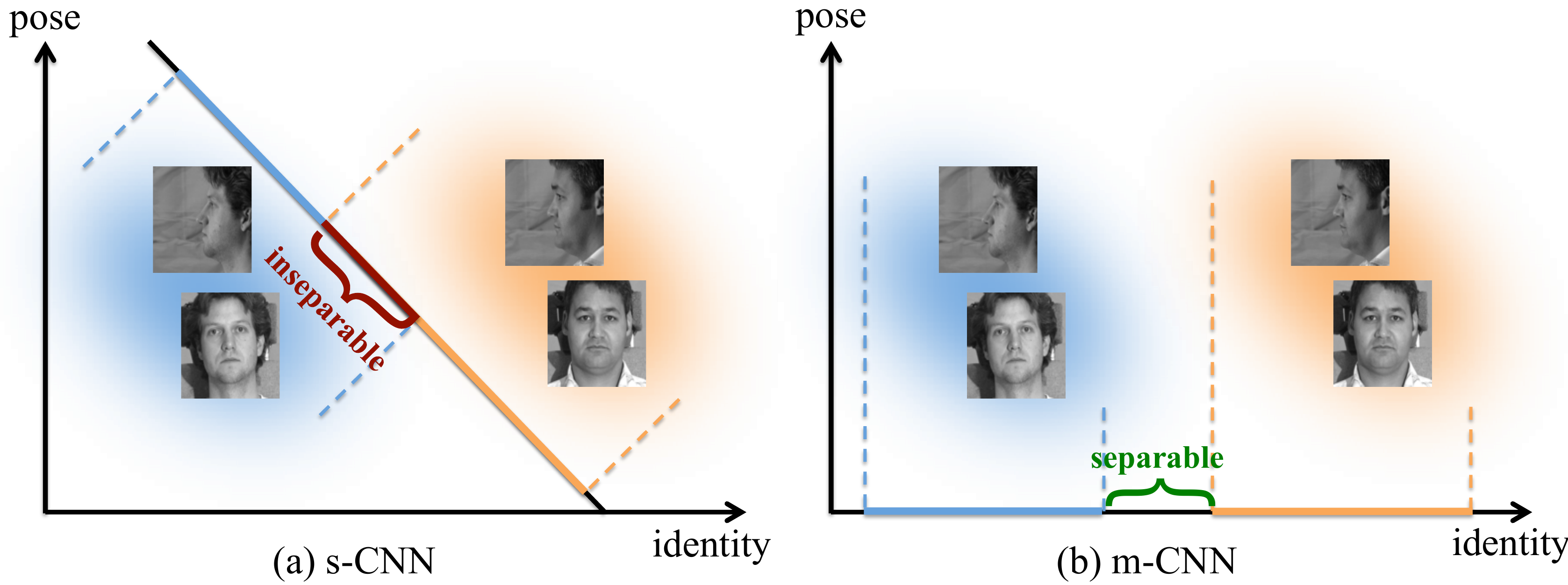}
\end{center}
\caption{We propose MTL to disentangle the PIE variations from learnt identity features. The above example shows two subjects at two different poses where identity is modeled in x-axis and pose in y-axis. (a) For single-task learning (s-CNN), the main variance is captured in $x$-$y$, resulting in an inseparable region between these two subjects. (b) For multi-task learning (m-CNN), identity is separable in $x$-axis by excluding $y$-axis that models the pose variation.}
\label{fig:concept}
\vspace{-4mm}
\end{figure}

We answer the questions of how and why PIE estimations can help face recognition by incorporating MTL into the CNN framework. 
We assume that different tasks share the same feature representation, which is learnt through several convolutional and pooling layers. 
A fully connected layer is added to the shared features for the classification of each task. 
We propose an energy-based weight analysis to explore how MTL works. 
And we observe that the side tasks serve as regularizations to learn a disentangled identity representation.
As shown in Figure~\ref{fig:concept}, when identity ($x$ axis) is mixed with pose variation ($y$ axis), single-task learning (s-CNN) for face recognition may learn a joint decision boundary along $x$-$y$, resulting in an inseparable region between different identities. 
In contrary, with multi-task learning (m-CNN), the shared feature space is learnt to model identity and pose separately. 
The identity features can exclude pose variation by selecting only the key dimensions that are essential for face recognition. 

One crucial problem in MTL is how to determine the importance of each task.
Prior work either treat different tasks equally~\cite{yim2015rotating} or obtain the weights by greedy search~\cite{tian2015pedestrian}.
We believe that our side tasks contribute differently to the main task of face recognition. 
However, it will be very time consuming or practically impossible to find the optimal weights for all side tasks via brute-force search. 
Instead, we propose a dynamic-weighting scheme where we only need to determine the overall weight for PIE estimations, and the CNN can learn to dynamically assign a loss weight to each side task during training. 
This is effective and efficient as will shown in Section~\ref{sec:exp}. 

Since pose variation is the most challenging one among other non-identity variations, and the proposed m-CNN already classifies all images into different pose groups, we propose to apply divide-and-conquer to CNN learning. 
Specifically, we develop a novel pose-directed multi-task CNN (p-CNN) where the pose labels can categorize the training data into three different pose groups, direct them through different routes in the network to learn pose-specific identity features in addition to the generic identity features. 
Similarly, the loss weights for extracting these two types of features are learnt dynamically in the CNN framework. 
During the testing stage, we propose a stochastic routing scheme to fuse the generic identity features and the pose-specific identity features for face recognition that is more robust to pose estimation errors. 
We find this technique to be very effective for pose-invariant face recognition especially for in-the-wild faces. 

This work utilizes {\it all} data in Multi-PIE~\cite{gross2010multi}, i.e., faces with the full range of PIE variations, as the main experimental dataset --- ideal for studying MTL for PIE-invariant face recognition. 
To the best of our knowledge, there is no prior face recognition work that studies the full range of variations in Multi-PIE. 
We also apply our method to in-the-wild datasets for pose-invariant face recognition. 
Since the ground truth label of the side task is unavailable, we use the estimated poses as labels for training. 

In summary, we make four contributions:
\begin{itemize}
\item we explore how and why PIE estimations can help face recognition;
\item we propose a dynamic-weighting scheme to learn the loss weights for different tasks automatically in the CNN framework;
\item we develop a pose-directed multi-task CNN to handle pose variation;
\item this is the most comprehensive and first face recognition study on entire Multi-PIE. We achieve comparable or superior performance to state-of-the-art methods on Multi-PIE, LFW~\cite{huang2007labeled}, CFP~\cite{sengupta2016frontal}, and IJB-A~\cite{klare2015pushing}. 
\end{itemize}

\section{Related Work}
\subsection{Face Recognition}
Face recognition is one of the most widely studied topics in computer vision. 
In this work, we study PIE-invariant face recognition on Multi-PIE and pose-invariant face recognition on in-the-wild datasets via CNN-based multi-task learning. 
Therefore, we focus our review on face recognition methods related to handling pose variation, MTL, and the usage of Multi-PIE dataset. 

\Paragraph{Pose-Invariant Face Recognition}
According to~\cite{ding2016comprehensive}, existing PIFR methods can be classified into four categories including: multi-view subspace learning~\cite{li2009maximizing,andrew2013deep}, pose-invariant feature extraction~\cite{chen2013blessing, schroff2015facenet}, face synthesis~\cite{zhu2015high, hassner2015effective}, and a hybrid approach of the above three~\cite{tran2017disentangled,yim2015rotating}. 
Our work belongs to the second category of extracting pose-invariant features. 
Some previous work in this category treat each pose separately by learning different models for face images with different poses. 
For example, Masi et al.~\cite{masi2016pose} propose pose-aware face recognition by learning a specific model for each type of face alignment and pose group. 
The idea of divide-and-conquer is similar to our work. 
Differently, we learn pose-invariant identity features for all poses jointly in one CNN framework. 
Xiong et al.~\cite{xiong2015conditional} propose a conditional CNN for face recognition, which can discover the modality information automatically during training. 
In contrast, we utilize the pose labels as a side task to better disentangle pose variation from the learnt identity features. 

\Paragraph{MTL for Face Recognition} 
For MTL-based face recognition methods, Ding et al.~\cite{ding2015multi} propose to transform the features of different poses into a discriminative subspace, and the transformations are learnt jointly for all poses with one task for each pose. 
\cite{yim2015rotating} develops a deep neural network to rotate a face image while preserving the identity, and the reconstruction of the face is considered as a side task, which has proved to be more effective than the single task model without appending the reconstruction layers. 
Similar work~\cite{zhu2014multi, tran2017disentangled} have developed along this direction to extract robust identity features and synthesize face images simultaneously. 
In this work, we treat face recognition as a multi-task problem with PIE estimations as the side tasks. 
It sounds intuitive to have PIE as side tasks for face recognition, but we are actually the first to consider this and we have found it to be very effective. 

\Paragraph{Multi-PIE}
Multi-PIE dataset consists of $754,200$ images of $337$ subjects with PIE variations. 
As a classic face dataset, it has been used to study face recognition robust to pose~\cite{li2012morphable, kan2014stacked}, illumination~\cite{han2010lighting, han2012separability}, and expression~\cite{chu20143d, zhu2015high}. 
Most prior work study the combined variations of pose and illumination~\cite{ding2015multi, zhang2013pose, zhu2014multi} with an increasing pose variation from half-profile~\cite{zhang2013pose} to a full range~\cite{xiong2015conditional}. 
Chu et al.~\cite{chu20143d} propose a framework for pose normalization and expression neutralization by using an extended 3D Morphable Model. 
The work of ~\cite{yang2013sparse} develops a sparse variation dictionary learning to study face recognition under a subset of PIE variations. 
However, only a very small subset is selected for experiments in~\cite{chu20143d, yang2013sparse}. 
To the best of our knowledge, this is the first effort using the entire set of Multi-PIE to study face recognition under a full span of PIE variations. 
A comparison of different experimental settings is shown in Table~\ref{tab_multipie}.

\subsection{Multi-Task Learning}
Multi-task learning has been widely studied in machine learning~\cite{gong2014efficient, argyriou2008convex}, natural language processing~\cite{collobert2008unified}, and computer vision~\cite{tian2015pedestrian, zhang2014facial} communities. 
In our work, we explore why and how MTL can help face recognition in a CNN-based framework. 
We focus our review on different regularizations in MTL, CNN-based MTL, and the problem of how to determine the importance of each task in MTL. 

\Paragraph{Regularizations}
The underlying assumption for most MTL algorithms is that different tasks are related to each other. 
Thus, a key problem is how to determine the task relatedness and take this into account to formulate the learning model. 
One common way is to learn a set of shared features for different tasks. 
The generalized linear model parameterizes each task with a weight vector. 
The weight vectors of all tasks form a weight matrix, which is regularized by $l_{2,1}$- norm~\cite{argyriou2008convex,obozinski2006multi} or trace norm~\cite{ji2009accelerated} to encourage a low-rank matrix. 
For example, Obozinski et al.~\cite{obozinski2006multi} propose to penalize the sum of $l_2$-norm of the blocks of weights associated with each feature across different tasks to encourage similar sparsity patterns. 
Lin et al.~\cite{lin2016multi} propose to learn higher order feature interaction without limiting to linear model for MTL. 
Other work~\cite{fei2013structured,zhang2012convex,lin2016interactive} propose to learn the task relationship from a task covariance matrix computed from the data.

\Paragraph{CNN-based MTL}
It is natural to fuse MTL with CNN to learn the shared features and the task-specific models.
For example,~\cite{zhang2014facial} proposes a deep CNN for joint face detection, pose estimation, and landmark localization.
Misra et. al.~\cite{misra2016cross} propose a cross-stitch network for MTL to learn the sharing strategy, which is difficult to scale to multiple tasks. 
Because it requires training one model for each task and introduces additional parameters in combining them. 
In~\cite{zhang2014improving}, a task-constrained deep network is developed for landmark detection with facial attribute classifications as the side tasks.
However, unlike the regularizations used in the MTL formulation in the machine learning community, there is no principled method to analysis how MTL works in the CNN framework. 
In this paper, we propose an energy-based weight analysis method to explore how MTL works. 
We discover that the side tasks of PIE estimations serve as regularizations to learn more discriminative identity features that are robust to PIE variations. 

\Paragraph{Importance of Each Task}
In MTL, it is important to determine the loss weights for different tasks.
The work of~\cite{yim2015rotating} uses equal weights for the tasks of face recognition and face frontalization. 
Tian et al.~\cite{tian2015pedestrian} propose to fix the weight for the main task to $1$, and obtain the weights of all side tasks via greedy search within $0$ and $1$. 
Let $t$ and $k$ be the number of side tasks and searched values respectively. 
This approach has two drawbacks. 
First, it is very inefficient as the computation scales to the number of tasks (complexity $tk$). 
Second, the optimal weight obtained for each task may not be jointly optimal. 
Further, the complexity would be $k^t$ if we search all combinations in a brute-force way. 
Zhang et al.~\cite{zhang2014facial} propose a task-wise early stopping to halt a task during training when the loss no longer reduces. 
However, a stopped task will never resume during training so the effect of this task may disappear. 
In contrast, we propose a dynamic-weighting scheme where we only determine the overall weight for all side tasks (complexity $k$) and let CNN learn to automatically distribute the weights to each side task. 
In this case when one task is saturated, we have observed the dynamic weights will reduce without the need to stop a specific task.

\section{The Proposed Approach}
In this section, we present the proposed approach by using Multi-PIE dataset as an example and extend it to in-the-wild datasets in the experiments. 
First, we propose a multi-task CNN (m-CNN) with dynamic weights for face recognition (the main task) and PIE estimations (the side tasks). 
Second, we propose a pose-directed multi-task CNN (p-CNN) to tackle pose variation by separating all poses into different groups and jointly learning pose-specific identity features for each group. 

\begin{figure*}[t!]
\begin{center}
\includegraphics[width=.98\textwidth]{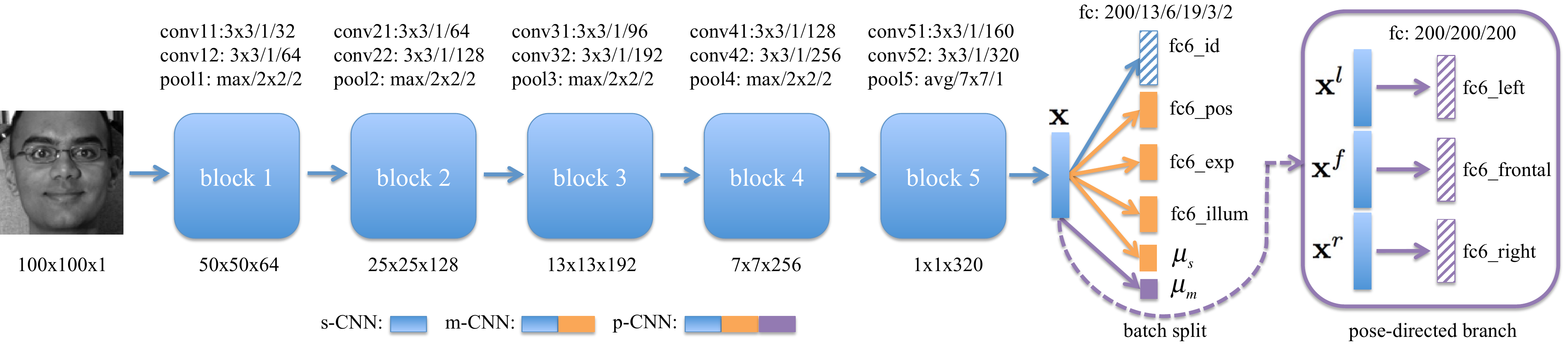}
\end{center}
\caption{The proposed m-CNN and p-CNN for face recognition. Each block reduces the spatial dimensions and increases the channels of the feature maps. The parameter format for the convolutional layer is: filter size / stride / filter number. The parameter format for the pooling layer is: method / filter size / stride. The feature dimensions after each block operation are shown on the bottom. The color indicates the component for each model. The dashed line represents the batch split operation as shown in Figure~\ref{fig:batch_split}. 
The layers with the stripe pattern are the identity features used in the testing stage for face recognition.}
\label{fig:overview}
\end{figure*}

\subsection{Multi-Task CNN}
We combine MTL with CNN framework by sharing some layers between different tasks. 
In this work, we adapt CASIA-Net~\cite{yi2014learning} with three modifications.
First, batch normalization (BN)~\cite{ioffe2015batch} is applied to accelerate the training process. 
Second, the contrastive loss is excluded to simplify our loss function.
Third, the dimension of the fully connected layer is changed according to different tasks. 
Details of the layer parameters are shown in Figure~\ref{fig:overview}. 
The network consists of five blocks each including two convolutional layers and a pooling layer. 
BN and ReLU~\cite{nair2010rectified} are used after each convolutional layer, which are omitted from the figure for clarity. 
Similar to~\cite{yi2014learning}, no ReLU is used after conv$52$ layer to learn a compact feature representation, and a dropout layer with a ratio of $0.4$ is applied after pool$5$ layer. 

Given a training set $\bf{D}$ with $N$ images and their labels: ${\bf{D}} = \{{\bf{I}}_i, {\bf{y}}_i\}_{i=1}^{N}$, where ${\bf{I}}_i$ is the image and ${\bf{y}}_i$ is a vector consisting of the identity label $y_i^d$ (main task) and the side task labels.
In our work, we consider three side tasks including pose ($y_i^p$), illumination ($y_i^l$), and expression ($y_i^e$).
We eliminate the sample index $i$ for clarity. 
As shown in Figure~\ref{fig:overview}, the proposed m-CNN extracts a high-level feature representation ${\bf{x}}\in \mathbb{R}^{D\times1}$: 
\begin{equation}
{\bf{x}} = f({\bf{I}}; {\bf{k}}, {\bf{b}}, {\boldsymbol{\gamma}}, {\boldsymbol{\beta}}),
\end{equation}
where $f(\cdot)$ represents the non-linear mapping from the input image to the shared features.
$\bf{k}$ and $\bf{b}$ are the sets of filters and bias of all convolutional layers. 
$\boldsymbol{\gamma}$ and $\boldsymbol{\beta}$ are the sets of scales and shifts in after BN layers~\cite{ioffe2015batch}. 
Let $\boldsymbol{\Theta} = \{{\bf{k}}, {\bf{b}}, \boldsymbol{\gamma}, \boldsymbol{\beta}\}$ denote all parameters to be learnt to extract the features $\bf{x}$. 

The extracted features ${\bf{x}}$, which is pool$5$ in our model, are shared among all tasks. 
Suppose ${\bf{W}}^d\in \mathbb{R}^{D\times D_d}$ and ${\bf{b}}^d\in \mathbb{R}^{D_d\times 1}$ are the weight matrix and bias vector in the fully connected layer for identity classification, where $D_d$ is the number of different identities in ${\bf{D}}$. 
The generalized linear model can be applied: 
\begin{equation}
{\bf{y}}^d = {{\bf{W}}^d}^\intercal {\bf{x}} + {\bf{b}}^d.
\end{equation}

${\bf{y}}^d$ is fed to a softmax layer to compute the probability of $\bf{x}$ belonging to each subject in the training set:
\begin{equation}
softmax({\bf{y}}^d)_n = p(\hat{y}^d=n|{\bf{x}}) = \frac{\exp({\bf{y}}^d_n)} {\sum_{j} \exp({\bf{y}}^d_j)}, 
\label{eqn:softmax}
\end{equation}
where ${\bf{y}}^d_j$ is the $j$th element in ${\bf{y}}^d$.
The $softmax(\cdot)$ function converts the output ${\bf{y}}^d$ to a probability distribution over all subjects and the subscript selects the $n$th element. 
Finally, the estimated identity $\hat{y}^d$ is obtained via: 
\begin{equation}
\hat{y}^d = \underset{n}{\mathrm{argmax}} \quad softmax({\bf{y}}^d)_n.
\end{equation}

Then the cross-entropy loss can be employed:
\begin{equation}
L({\bf{I}}, y^d) = - \log(p(\hat{y}^d = y^d|{\bf{I}}, \boldsymbol{\Theta}, {\bf{W}}^d, {\bf{b}}^d)).
\end{equation}

Similarly, we formulate the losses for the side tasks. 
Let ${\bf{W}} = \{ {\bf{W}}^d, {\bf{W}}^p, {\bf{W}}^l, {\bf{W}}^e \}$ represent the weight matrices for identity and PIE classifications. 
The bias terms are eliminated for simplicity. 
Given the training set ${\bf{D}}$, our m-CNN aims to minimize the combined loss of all tasks:
\begin{equation}
\small
\begin{split}
\underset{\boldsymbol{\Theta}, {\bf{W}}} {\mathrm{argmin}} \quad \alpha_d \sum_{i=1}^{N} L({\bf{I}}_i, y^d_i) + \alpha_p \sum_{i=1}^{N} L({\bf{I}}_i, y^p_i) + \\
\alpha_l \sum_{i=1}^{N} L({\bf{I}}_i, y^l_i) + \alpha_e \sum_{i=1}^{N} L({\bf{I}}_i, y^e_i),
\end{split}
\label{eqn:obj}
\end{equation}
where $\alpha_d$, $\alpha_p$, $\alpha_l$, $\alpha_e$ control the importance of each task. 
It becomes a single-task model (s-CNN) when $\alpha_{p,l,e} = 0$. 
The loss drives the model to learn both the parameters $\boldsymbol{\Theta}$ for extracting the shared features and $\bf{W}$ for the classification tasks. 
In the testing stage, the features before the softmax layer (${\bf{y}}^d$) are used for face recognition by applying a face matching procedure based on cosine similarity. 

\subsection{Dynamic-Weighting Scheme}
\label{sec:dynamic}
In CNN-based MTL, it is an open question on how to set the loss weight for each task. 
Prior work either treat all tasks equally~\cite{yim2015rotating} or obtain weights via brute-force search~\cite{tian2015pedestrian}. 
However neither works in our case. 
First, we believe that PIE estimations should contribute differently to our main task.
Second, it is very time-consuming to search for all weight combinations especially considering the training time for CNN models. 
To solve this problem, we propose a dynamic-weighting scheme to automatically assign the loss weights to each side task during training. 

First, we set the weight for the main task to $1$, i.e.~$\alpha_d=1$.
Second, instead of finding the loss weight for each task, we find the summed loss weight for all side tasks, i.e.~$\varphi_s = \alpha_p + \alpha_l + \alpha_e$, via brute-force search in a validation set.
Our m-CNN learns to allocate $\varphi_s$ to three side tasks. 
As shown in Figure~\ref{fig:overview}, we add a fully connected layer and a softmax layer to the shared features $\bf{x}$ to learn the dynamic weights. 
Let $\boldsymbol{\omega}_s\in \mathbb{R}^{D\times 3}$ and $\boldsymbol{\epsilon}_s \in \mathbb{R}^{3\times 1}$ denote the weight matrix and bias vector in the fully connected layer, 
\begin{equation}
{\boldsymbol{\mu}_s} = softmax({\boldsymbol{\omega}_s}^\intercal {\bf{x}} + \boldsymbol{\epsilon}_s),
\end{equation}
where $\boldsymbol{\mu}_s = [\mu_p, \mu_l, \mu_e]^\intercal $ are the dynamic weight percentages for the side tasks with $\mu_p+\mu_l+\mu_e=1$.
The function $softmax$ converts the dynamic weights to positive values that sum to $1$. 
So Equation~\ref{eqn:obj} becomes:
\begin{equation}
\small
\begin{split}
\underset{\boldsymbol{\Theta}, {\bf{W}}, \boldsymbol{\omega}_s} {\mathrm{argmin}} \quad \sum_{i=1}^{N} L({\bf{I}}_i, y^d_i) + \varphi_s \Big[ \mu_p \sum_{i=1}^{N} L({\bf{I}}_i, y^p_i) + \\
\mu_l \sum_{i=1}^{N} L({\bf{I}}_i, y^l_i) + \mu_e \sum_{i=1}^{N} L({\bf{I}}_i, y^e_i) \Big] \\
s.t. \quad \mu_p+\mu_l+\mu_e=1, \quad \quad \quad \quad \quad
\end{split}
\label{eqn:obj1}
\end{equation}

The multiplications of the overall loss weight $\varphi_s$ with the learnt dynamic percentage $\mu_{p,l,e}$ are the dynamic loss weights for each side task.

We use mini-batch Stochastic Gradient Descent (SGD) to solve the above optimization problem where the dynamic weights are averaged over a batch of samples. 
Intuitively, we expect the dynamic-weighting scheme to behave in two different aspects in order to minimize the loss in Equation~\ref{eqn:obj1}. 
First, since our main task contribute mostly to the final loss ($\varphi_s <1$), the side task with the largest contribution to the main task should have the highest weight in order to reduce the loss of the main task. 
Second, our m-CNN should assign a higher weight for an easier task with a lower loss so as to reduce the overall loss. 
We have observed these effects as will shown in the experiments.

\subsection{Pose-Directed Multi-Task CNN}
It is very challenging to learn a non-linear mapping to estimate the correct identity from a face image with arbitrary PIE, given the diverse variations in the data. 
This challenge has been encountered in classic pattern recognition work.
For example, in order to handle pose variation, ~\cite{li2006bagging} proposes to construct several face detectors where each of them is in charge of one specific view. 
Such a divide-and-conquer scheme can be applied to CNN learning because the side tasks can ``divide'' the data and allow the CNN to better ``conquer'' them by learning tailored mapping functions.

Therefore, we propose a novel task-directed multi-task CNN where the side task labels categorizes the training data into multiple groups, and directs them to different routes in the network.
Since pose is considered as the primary challenge in face recognition~\cite{xiong2015conditional, zhang2013pose, zhu2014multi}, we propose pose-directed multi-task CNN (p-CNN) to handle pose variation. 
However, it is applicable to any other variation. 

\begin{figure}[t!]
\begin{center}
\includegraphics[width=0.48\textwidth]{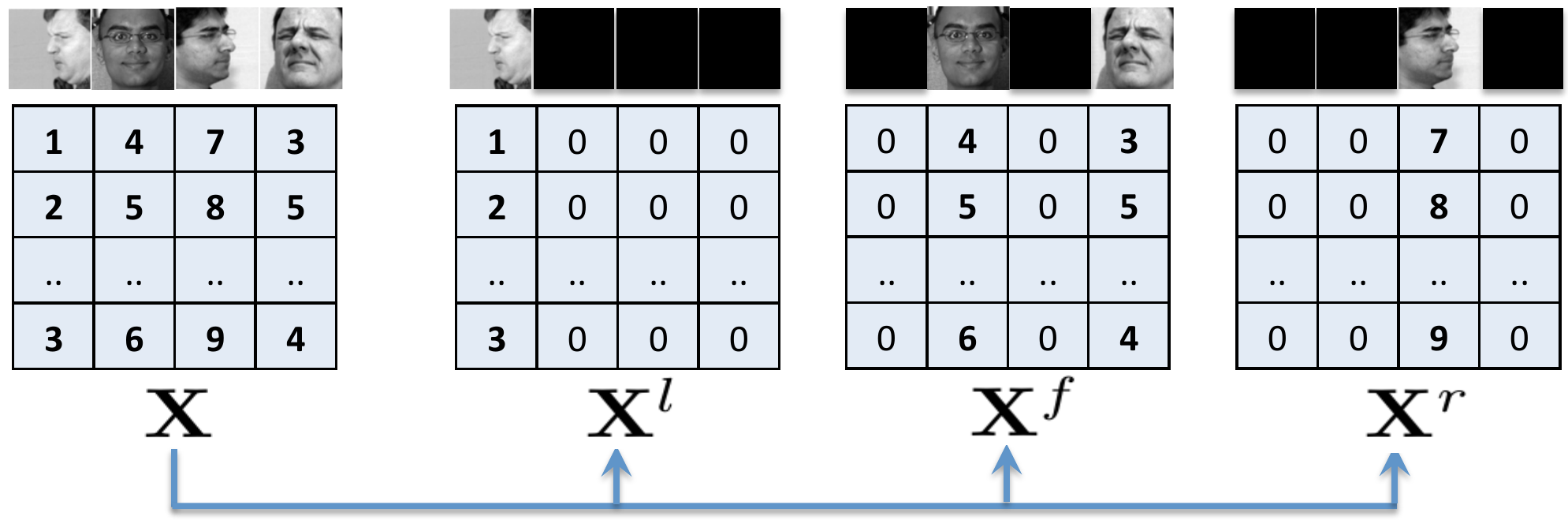}
\end{center}
\caption{Illustration of the batch split operation in p-CNN. The first row shows the input images and the second row shows a matrix representing the features $\bf{x}$ for each sample. After batch split, one batch of samples is separated into three batches with each only consists of the samples belonging to a specific pose group.}
\label{fig:batch_split}
\end{figure}

\begin{figure}[t!]
\begin{center}
\includegraphics[width=0.45\textwidth]{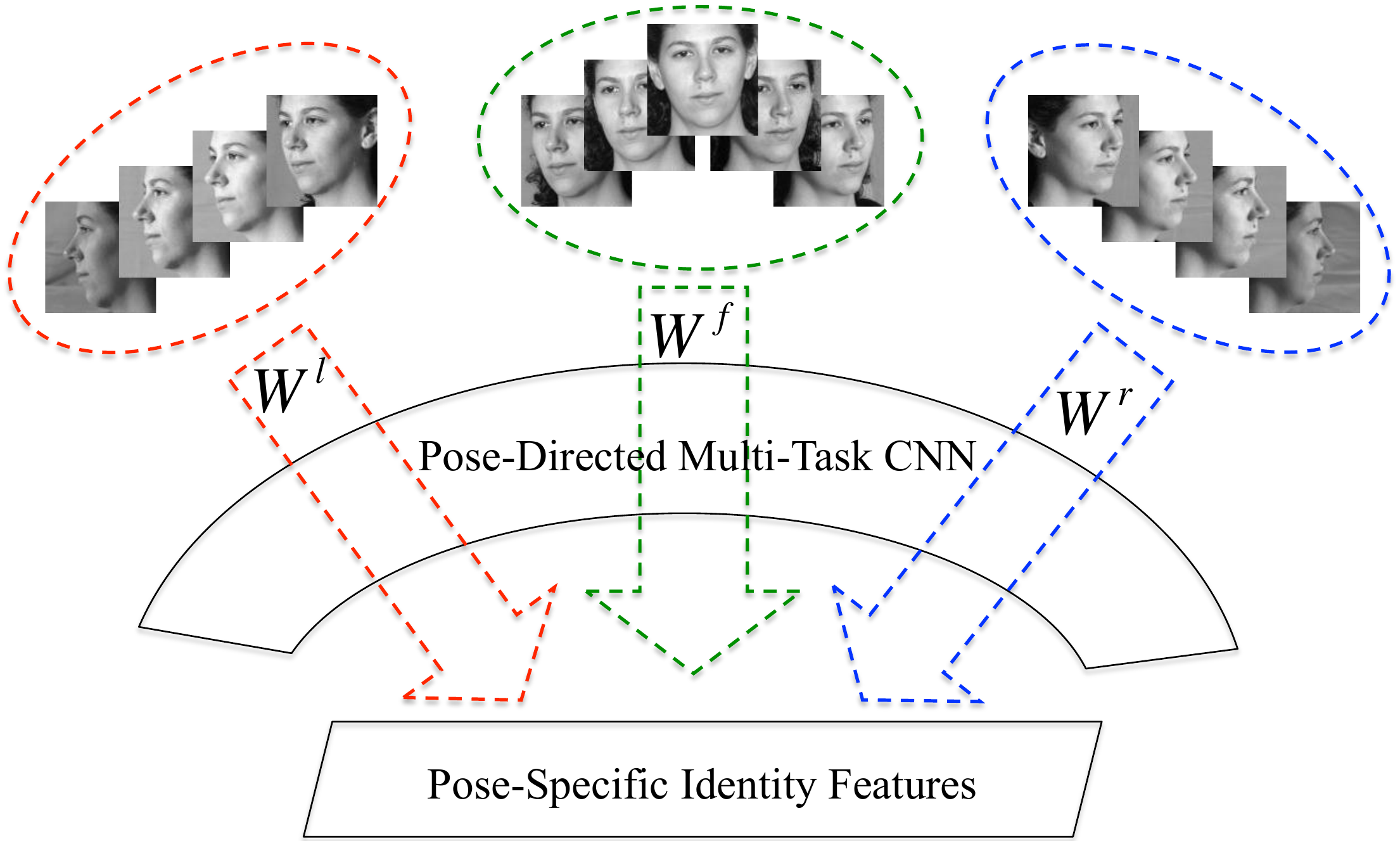}
\end{center}
\caption{The proposed pose-directed multi-task CNN aims to learn pose-specific identity features jointly for all pose groups.}
\label{fig:pcnn}
\end{figure}

As shown in Figure~\ref{fig:overview}, p-CNN is built on top of m-CNN by adding the pose-directed branch (PDB). 
The PDB groups face images with similar poses to learn pose-specific identity features via a batch split operation.
We separate the training set into three groups according to the pose labels: left profile ($G^l$), frontal ($G^f$), and right profile ($G^r$). 
As shown in Figure~\ref{fig:batch_split}, the goal of batch split is to separate a batch of $N_0$ samples (${\bf{X}} = \{ {\bf{x}}_i\}_{i=1}^{N_0}$) into three batches ${\bf{X}}^l$, ${\bf{X}}^f$, and ${\bf{X}}^r$, which are of the same size as ${\bf{X}}$.
During training, the ground truth pose is used to assign a face image into the correct group. 
Let us take the frontal group as an example:
\begin{equation}
\small
{\bf{X}}^f_i = 
\begin{cases}
{\bf{x}}_i,& \text{if } y_i^p\in G^f\\
{\bf{0}}, & \text{otherwise},
\end{cases}
\end{equation}
where $\bf{0}$ denotes a vector of all zeros with the same dimension as ${\bf{x}}_i$. 
The assignment of $\bf{0}$ is to avoid the case when no sample is passed into one group, the next layer will still have valid input. 
Therefore, ${\bf{X}}$ is separated into three batches where each batch consists of only the samples belonging to the corresponding pose group. 
Each group learns a pose-specific mapping to a joint space, resulting in three different sets of weights: $\{ {\bf{W}}^l, {\bf{W}}^f, {\bf{W}}^r \}$. 
This process is illustrated in Figure~\ref{fig:pcnn}. 

Our p-CNN aims to learn two types of identity features: 
${\bf{W}}^{d}$ is the weight matrix to extract the generic identity features that is robust to all poses;
${\bf{W}}^{l,f,r}$ are the weight matrices to extract the pose-specific identity features that are robust within a small pose range. 
Both tasks are considered as our main tasks. 
Similar to the dynamic-weighting scheme in m-CNN, we use dynamic weights to combine our main tasks as well.
The summed loss weight for these two tasks is $\varphi_m = \alpha_d + \alpha_g$. 
Let $\boldsymbol{\omega}_m\in \mathbb{R}^{D\times 2}$ and $\boldsymbol{\epsilon}_m \in \mathbb{R}^{2\times 1}$ denote the weight matrix and bias vector for learning the dynamic weights, 
\begin{equation}
{\boldsymbol{\mu}_m} = softmax({\boldsymbol{\omega}_m}^\intercal {\bf{x}} + \boldsymbol{\epsilon}_m).
\end{equation}

We have $\boldsymbol{\mu}_m = [\mu_d, \mu_g]^\intercal$ as the dynamic weights for generic identity classification and pose-specific identity classification. 
Finally, the loss of p-CNN is formulated as:
\begin{equation}
\small
\begin{split}
\underset{\boldsymbol{\Theta}, {\bf{W}}, \boldsymbol{\omega}} {\mathrm{argmin}} \quad \varphi_m \Big[\mu_d \sum_{i=1}^{N} L({\bf{I}}_i, y^d_i) + \mu_g \sum_{g=1}^{G}\sum_{i=1}^{N_g} L({\bf{I}}_i, y^d_i)\Big] + \\
\varphi_s\Big[ \mu_p \sum_{i=1}^{N} L({\bf{I}}_i, y^p_i) + \mu_l \sum_{i=1}^{N} L({\bf{I}}_i, y^l_i) + \mu_e \sum_{i=1}^{N} L({\bf{I}}_i, y^e_i) \Big] \\
s.t. \quad \mu_d + \mu_g = 1, \quad \mu_p+\mu_l+\mu_e=1, \quad \quad \quad \quad
\end{split}
\label{eqn:obj2}
\end{equation}
\normalsize
where $G=3$ is the number of pose groups and $N_g$ is the number of training images in the $g$-th group. 
$\boldsymbol{\omega} =\{ \boldsymbol{\omega}_m, \boldsymbol{\omega}_s\}$ is the set of parameters to learn the dynamic weights for both the main and side tasks. 
We set $\varphi_m = 1$. 

\begin{figure}[t]
\begin{center}
\includegraphics[width=0.4\textwidth]{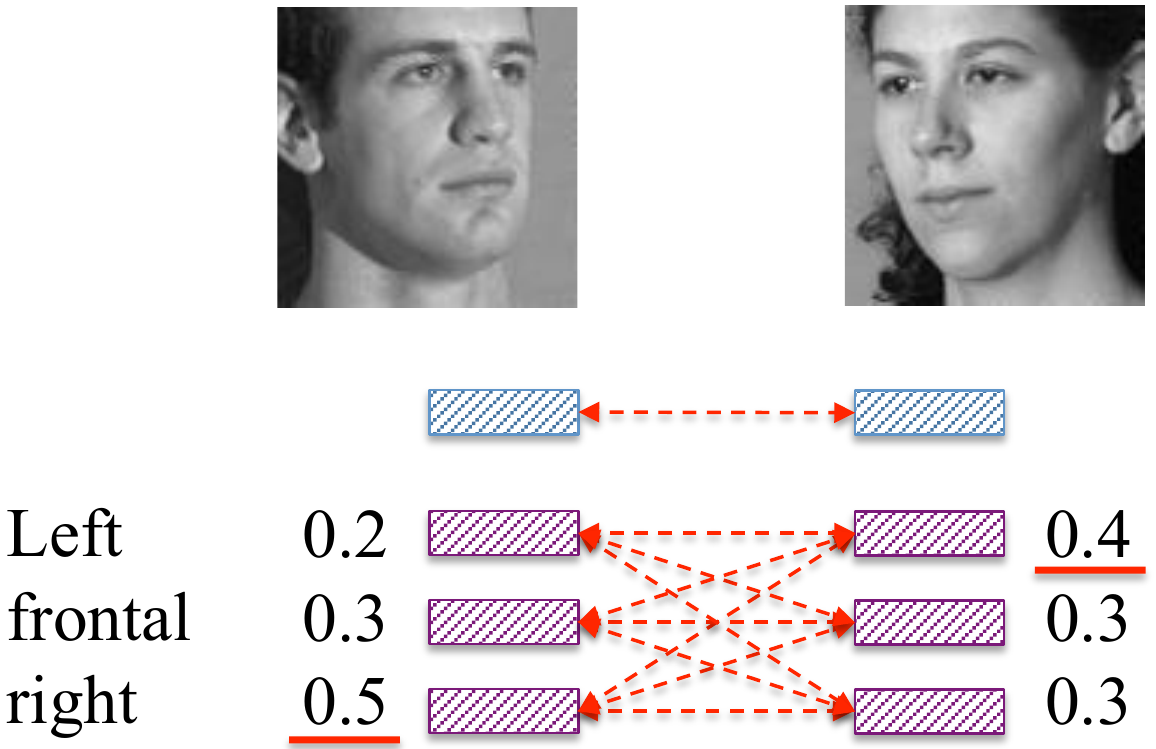}
\end{center}
\caption{Blue bars are the generic identity features and purple bars are the pose-specific features. The numbers are the probabilities of each input image belonging to each pose group. The proposed stochastic routing in the testing stage taking account of all pair comparisons so that it is more robust to pose estimation errors.}
\label{fig:stochastic}
\end{figure}

\begin{table*}[t!]
\small
\begin{center}
\caption{Performance comparison ($\%$) of single-task learning (s-CNN), multi-task learning (m-CNN) with its variants, and pose-directed multi-task learning (p-CNN) on the entire Multi-PIE dataset. }
\label{tab:multitask}
\begin{tabular}{ lccccc}
\hline 
model & loss weights & rank-1 (all / left / frontal /right) & pose & illum & exp \\ \hline \hline
s-CNN & $\alpha_d/\alpha_p/\alpha_l/\alpha_e=1$ & $75.67$ / $71.51$ / $82.21$ / $73.29$ & $99.87$ & $96.43$ & $92.44$ \\ \hline
m-CNN: id+pos & $\alpha_d=1, \alpha_p=0.1$ & $78.06$ / $75.06$ / $82.91$ / $76.21$ & $99.78$ & \textendash & \textendash \\ 
m-CNN: id+illum & $\alpha_d=1, \alpha_l=0.1$ & $77.30$ / $74.87$ / $82.83$ / $74.21$ & \textendash & $93.57$ & \textendash \\ 
m-CNN: id+exp & $\alpha_d=1, \alpha_e=0.1$ & $77.76$ / $75.48$ / $82.32$ / $75.48$ & \textendash & \textendash & $90.93$ \\ 
m-CNN: id+all & $\alpha_d=1, \alpha_{p,l,e}=0.033$ & $77.59$ / $74.75$ / $82.99$ / $75.04$ & $99.75$ & $88.46$ & $79.97$ \\ 
m-CNN: id+all (dynamic) & $\alpha_d=1, \varphi_s=0.1$ & $79.35$ / $76.60$ / $84.65$ / $76.82$ & $99.81$ & $93.40$ & $91.47$ \\ \hline
p-CNN & $\varphi_m = 1, \varphi_s = 0.1$ & $79.55$ / $76.14$ / $84.87$ / $77.65$ & $99.80$ & $90.58$ & $90.02$ \\ \hline\hline
\end{tabular}
\end{center}
\end{table*}

\Paragraph{Stochastic Routing} Given a face image in the testing stage, we can extract the generic identity features (${\bf{y}}^d$), the pose-specific identity features ($\{{\bf{y}}^g\}_{g=1}^3$), as well as estimate the probabilities ($\{p^g\}_{g=1}^3$) of the input image belonging to each pose group by aggregating the probabilities from the pose classification side task. 
As shown in Figure~\ref{fig:stochastic}, for face matching, we can compute the distance of the generic identity features and the distance of the pose-specific identity features by selecting the pose group with the largest probability (red underline). 
However, the pose estimation error may cause inferior feature extraction results, which is inevitable especially for unconstrained faces.

To solve this problem, we propose a stochastic routing scheme by taking account of all comparisons weighted by the probabilities. 
Specifically, the distance $c$ between a pair of face images ($\mathbf{I}_1$ and $\mathbf{I}_2$) is computed as the average between the distance of the generic identity features (${\bf{y}}_1^d$, ${\bf{y}}_2^d$) and weighted distance of the pose-specific identity features ($\{{\bf{y}}_1^g$\}, \{${\bf{y}}_2^g$\}): 
\begin{equation}
c = \frac{1}{2}h({\bf{y}}^d_1, {\bf{y}}^d_2) + \frac{1}{2} \sum_{i=1}^{3}\sum_{j=1}^{3} h({\bf{y}}^i_1, {\bf{y}}^j_2)\cdot p^i_1 \cdot p^j_2,
\end{equation}
where $h(\cdot)$ is the cosine distance metric used to measure the distance between two feature vectors. 
The proposed stochastic routing accounts for all combinations of the pose-specific identity features weighted by the probabilities of each combination.
We treat the generic features and pose-specific features equally and fuse them for face recognition.

\section{Experiments}
\label{sec:exp}
We evaluate the proposed m-CNN and p-CNN under two settings:
(1) face identification on Multi-PIE with PIE estimations being the side tasks;
(2) face verification/identification on in-the-wild datasets including LFW, CFP, and IJB-A, where pose estimation is the only side task. 
Further, we analysis the effect of MTL on Multi-PIE and discover that the side tasks regularize the network to learn a disentangled identity representation for PIE-invariant face recognition. 

\subsection{Face Identification on Multi-PIE}
\label{exp:multipie}
\Paragraph{Experimental Settings}
Multi-PIE dataset consists of $754,200$ images of $337$ subjects recorded in $4$ sessions. 
Each subject was recorded with $15$ different cameras where $13$ at the head height spaced at $15^\circ$ interval and $2$ above the head to simulate a surveillance camera view. 
For each camera, a subject was imaged under $19$ different illuminations. 
In each session, a subject was captured with $2$ or $3$ expressions, resulting in a total of $6$ different expressions across all sessions. 
In our work, we use the entire dataset including all PIE variations. 
For the two cameras above the head, their poses are labeled as $\pm45^\circ$. 
The first $200$ subjects are used for training. 
The remaining $137$ subjects are used for testing, where one image with frontal pose, neutral illumination, and neutral expression for each subject is selected as the gallery set and the remaining as the probe set.

We use the landmark annotations provided in~\cite{el2013scalable} to align each face to a canonical view of size $100\times100$.
The images are normalized by subtracting $127.5$ and dividing by $128$, similar to~\cite{wen2016discriminative}.
We use Caffe~\cite{jia2014caffe} with our modifications. 
The momentum is set to $0.9$ and the weight decay to $0.0005$. 
All models are trained for $20$ epochs from scratch with a batch size of $4$.
The learning rate starts at $0.01$ and reduces at $10$th, $15$th, and $19$th epochs with a factor of $0.1$. 
The features before the softmax layer are used for face matching based on cosine similarity. 
The rank-$1$ identification rate is reported as the face recognition performance. 
For the side tasks, the mean accuracy over all classes is reported. 

For m-CNN model training, we randomly select a subset of $20$ subjects from the training set to formulate a validation set to find the optimal loss weight for all side tasks. 
We obtain $\varphi_s = 0.1$ via brute-force search.
For p-CNN model training, we split the training set into three groups based on the yaw angle of the image: right profile ($-90^\circ, -75^\circ, -60^\circ$, $-45^\circ$), frontal ($-30^\circ$, $-15^\circ, 0^\circ, 15^\circ$, $30^\circ$), and left profile ($45^\circ$, $60^\circ, 75^\circ, 90^\circ$). 

\begin{figure*}[t!]
\centering
\begin{tabular}{@{}cccc@{}}
\includegraphics[trim=0 0 0 0, clip, width=.235\textwidth]{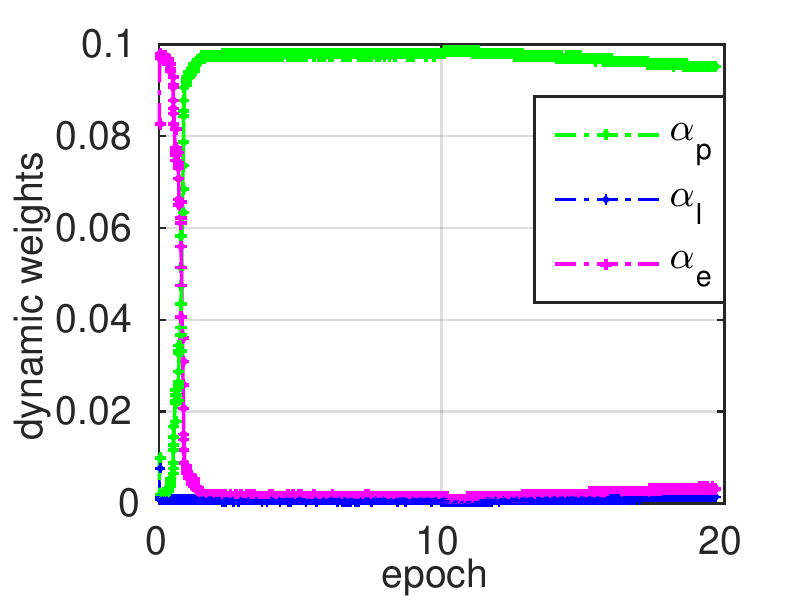} &
\includegraphics[trim=0 0 0 0, clip, width=.235\textwidth]{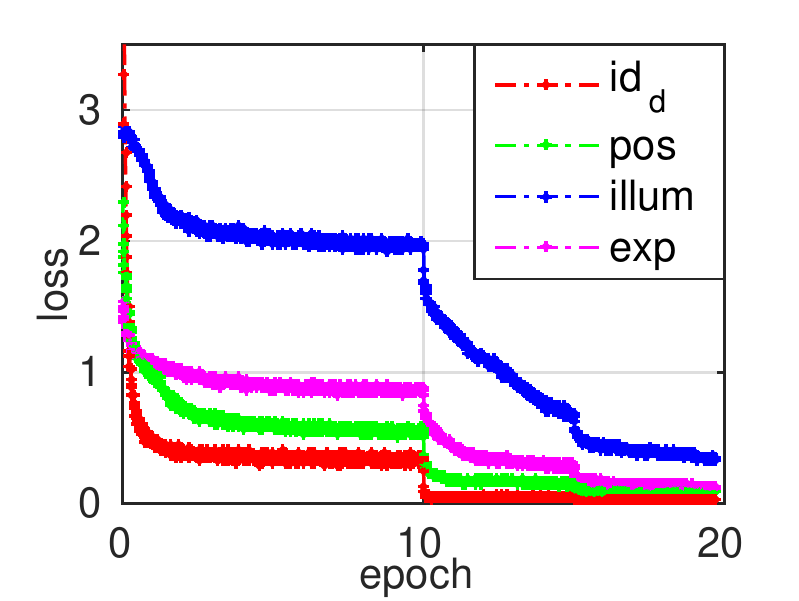} &
\includegraphics[trim=0 0 0 0, clip, width=.235\textwidth]{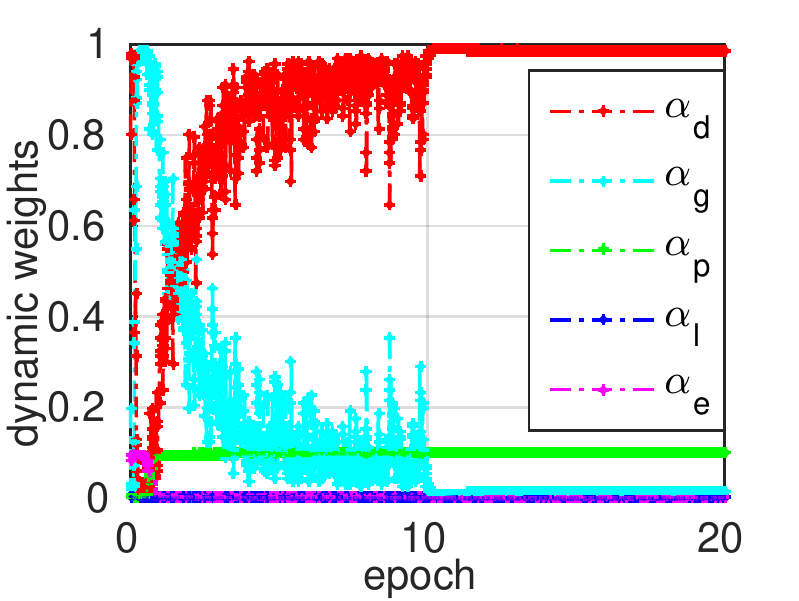} &
\includegraphics[trim=0 0 0 0, clip, width=.235\textwidth]{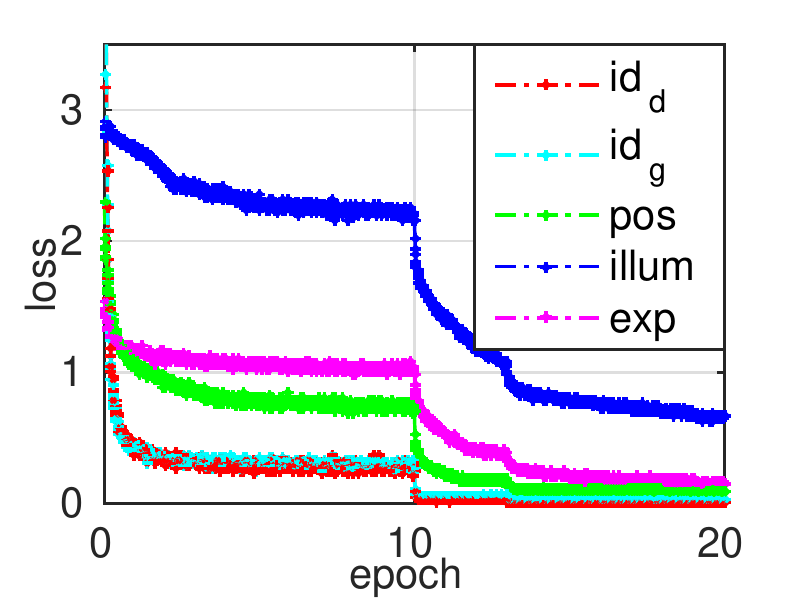} \\ [-1mm]
\footnotesize (a) m-CNN: dynamic weights & \footnotesize (b) m-CNN: losses & \footnotesize (c) p-CNN: dynamic weights & \footnotesize (d) p-CNN: losses\\
\end{tabular}
\caption{The learnt dynamic weights and the losses of each task for m-CNN and p-CNN models during the training process.}
\label{fig:dynamic_weights} 
\end{figure*}

\begin{table*}[t]
\begin{center}
\caption{Multi-PIE performance comparison in setting \RNum{3} of Table~\ref{tab_multipie}.} 
\label{tab:compare_1}
\begin{tabular}{lccccccc}
\hline 
& Avg. & $\pm90^\circ$ & $\pm75^\circ$ & $\pm60^\circ$ & $\pm45^\circ$ & $\pm30^\circ$ & $\pm15^\circ$ \\ \hline\hline
Fisher Vector~\cite{simonyan2013fisher} & 66.60 & 24.53 & 45.51 & 68.71 & 80.33 & 87.21 & 93.30 \\
FIP$\_20$~\cite{zhu2013deep} & $67.87$ & $34.13$ & $47.32$ & $61.64$ & $78.89$ & $89.23$ & $95.88$ \\ 
FIP$\_40$~\cite{zhu2013deep} & $70.90$ & $31.37$ & $49.10$ & $69.75$ & $85.54$ & $92.98$ & $96.30$ \\ 
c-CNN~\cite{xiong2015conditional} & $73.54$ & $41.71$ & $55.64$ & $70.49$ & $85.09$ & $92.66$ & $95.64$ \\ 
c-CNN Forest~\cite{xiong2015conditional} & $76.89$ & $47.26$ & $60.66$ & $74.38$ & $89.02$ & $94.05$ & $96.97$ \\ \hline
s-CNN (ours) & $88.45$ & $76.72$ & $82.80$ & $88.71$ & $85.18$ & $96.89$ & $98.41$ \\ 
m-CNN (ours) & $90.08$ & $76.72$ & $84.97$ & $89.75$ & $89.15$ & $97.40$ & $99.02$ \\ 
p-CNN (ours) & ${\bf{91.27}}$ & ${\bf{76.96}}$ & ${\bf{87.83}}$ & ${\bf{92.07}}$ & $\bf{90.34}$ & $\bf{98.01}$ & ${\bf{99.19}}$ \\ \hline\hline
\end{tabular}
\end{center}
\end{table*}

\Paragraph{Effects of MTL}
Table~\ref{tab:multitask} shows the performance comparison of single-task learning (s-CNN), multi-task learning (m-CNN), and pose-directed multi-task learning (p-CNN) on the entire Multi-PIE dataset. 
First, we train four single-task models for identity (id), pose (pos), illumination (illum), and expression (exp) classification respectively. 
As shown in the first row of Table~\ref{tab:multitask}, the rank-$1$ identification rate of s-CNN is only $75.67\%$.
The performance of the frontal pose group is much higher than those of the profile pose groups, indicating that pose variation is indeed a big challenge for face recognition. 
Among all side tasks, pose estimation is the easiest, followed by illumination, and expression is the most difficult task. 
This is caused by two potential reasons: 1) discriminating expression is more challenging due to the non-rigid face deformation;
2) the data distribution over different expressions is unbalanced with insufficient training data for some expressions. 

Second, we train multiple m-CNN models by adding only one side task at a time in order to evaluate the influence of each side task. 
We use ``id+pos'', ``id+illum'', and ``id+exp'' to represent these variants and compare them to the performance of adding all side tasks denoted as ``id+all''.
To evaluate the effects of the dynamic-weighting scheme, we train a model with fixed loss weights for the side tasks as: $\alpha_p=\alpha_l=\alpha_e=\varphi_s/3=0.033$.
The summation of the loss weights for all side tasks are equal to $\varphi_s$ for all m-CNN variants in Table~\ref{tab:multitask} for a fair comparison. 

Comparing the rank-$1$ identification rates of s-CNN and m-CNNs, it is obvious that adding the side tasks is always helpful for the main task. 
The improvement of face recognition is mostly on the profile faces with MTL. 
The m-CNN ``id+all'' with dynamic weights shows superior performance to others not only in rank-$1$ identification rate, but also in the side task estimations. 
Further, the lower rank-$1$ identification rate of ``id+all'' w.r.t~``id+pos'' indicates that more side tasks do not necessarily lead to better performance without properly setting the loss weights.
In contrast, the proposed dynamic-weighting scheme effectively improves the performance to $79.35\%$ from the fixed weighting of $77.59\%$. 

Third, we train the p-CNN by adding the PDB to m-CNN ``id+all'' with dynamic weights. 
The loss weight for the main task is $\varphi_m=1$ and $\varphi_s=0.1$ for the side tasks. 
The proposed dynamic-weighting scheme allocates the loss weight to both two main tasks and three side tasks. 
Our p-CNN aims to learn generic identity features and pose-specific identity features, which are fused for face recognition. 
As shown in the last row of Table~\ref{tab:multitask}, p-CNN further improves the rank-$1$ identification rate to $79.55\%$. 

\Paragraph{Dynamic-Weighting Scheme} 
Figure~\ref{fig:dynamic_weights} shows the dynamic weights and losses during training for m-CNN and p-CNN. 
For m-CNN, the expression classification task has the largest weight in the first epoch as it has the highest chance to be correct with random guess due to the fact that it has the least number of classes. 
As training goes on, pose classification takes over because it is the easiest task (highest accuracy in s-CNN) and also most helpful for face recognition (compare id+pos to id+exp and id+illum). 
$\alpha_p$ starts to decrease at $11$th epoch when pose classification is almost saturated. 
The increased $\alpha_l$ and $\alpha_e$ lead the reduction in the losses of expression and illumination classifications. 
As we expected, the dynamic-weighting scheme assigns a higher loss weight for the easiest and/or the most helpful side task. 

For p-CNN, the loss weights and losses for the side tasks behave similarly to those of m-CNN. 
For the two main tasks, the dynamic-weighting scheme assigns a higher loss weight to the easier task at the moment. 
At the beginning, learning the pose-specific identity features is an easier task than learning the generic identity features. 
Therefore the loss weight $\alpha_g$ is higher than $\alpha_d$. 
As training goes on, $\alpha_d$ increases as it has a lower loss. 
Their losses reduce in a similar way, i.e., the error reduction in one task will also contribute to the other. 

\begin{figure*}[t!]
\centering
\begin{tabular}{@{}ccc@{}}
\includegraphics[trim=0 0 0 0, clip, width=.31\textwidth]{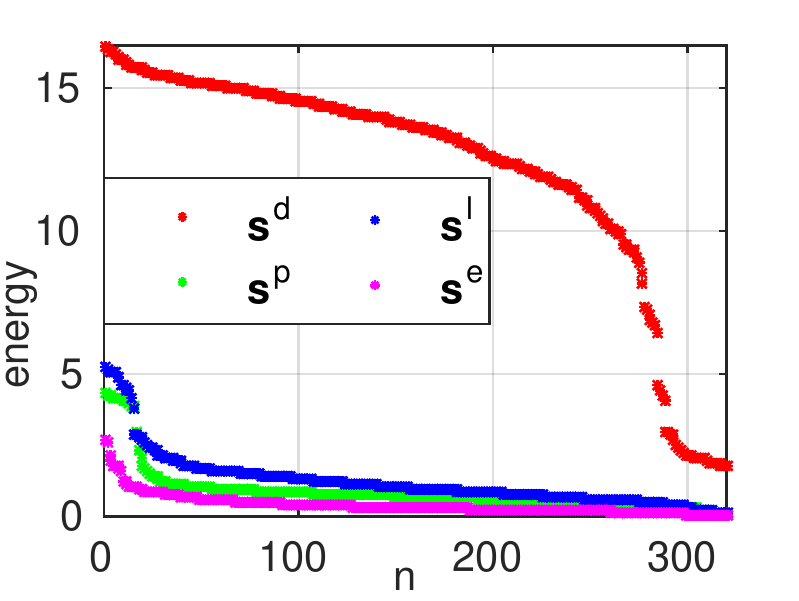} & 
\includegraphics[trim=0 0 0 0, clip, width=.33\textwidth]{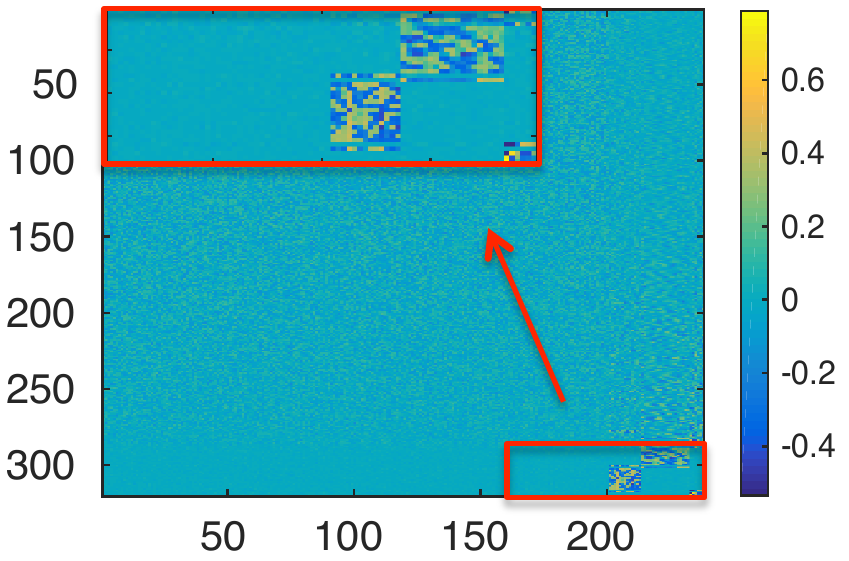} &
\includegraphics[trim=0 0 0 0, clip, width=.31\textwidth]{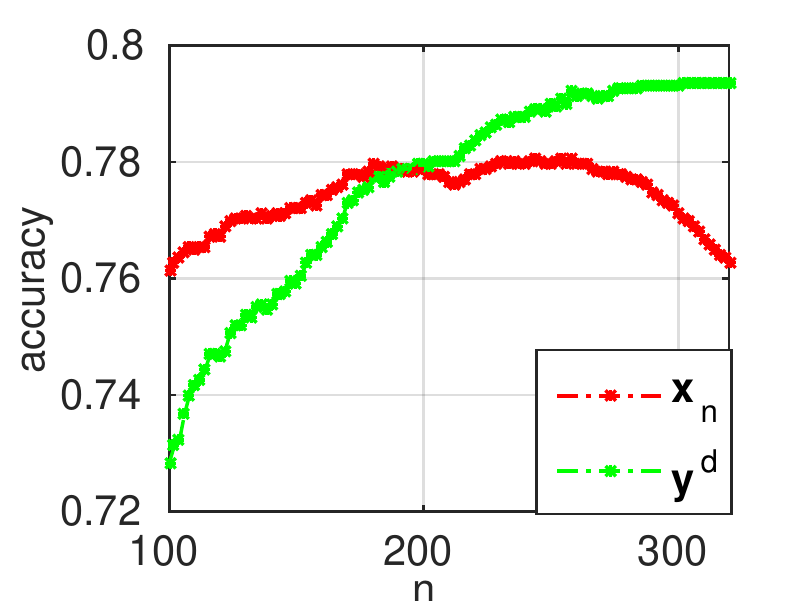} \\ 
\small (a) & \small (b) & \small (c) \\
\end{tabular}
\caption{Analysis on the effects of MTL: (a) the sorted energy vectors for all tasks; (b) visualization of the weight matrix ${\bf{W}}^{all}$ where the red box in the top-left is a zoom-in view of the bottom-right; (c) the face recognition performance with varying feature dimensions.}
\label{fig:mCNN} 
\end{figure*}

\Paragraph{Compare to Other Methods}
As shown in Table~\ref{tab_multipie}, no prior work uses the entire Multi-PIE for face recognition. 
To compare with state of the art, we choose to use setting \RNum{3} to evaluate our method since it is the most challenging setting with all poses included. 
The network structures and parameter settings are kept the same as those of the full set except that the outputs of the last fully connected layers are changed according to the number of classes for each task.
And only pose and illumination are considered as the side tasks. 

The performance are shown in Table~\ref{tab:compare_1}.
Our s-CNN already outperforms c-CNN forest~\cite{xiong2015conditional}, which is an ensemble of three c-CNN models. 
This is attributed to the deep structure of CASIA-Net~\cite{yi2014learning} compared to~\cite{xiong2015conditional}.
Moreover, our m-CNN and p-CNN further outperform s-CNN with significant margins, especially for non-frontal faces. 
We want to stress the improvement margin between our method $91.27\%$ and the prior work of $76.89\%$ --- a relative error reduction of $62\%$.
This huge margin is rarely seen in prior face recognition work, especially on a classic benchmark dataset, which is a testimony on our contribution to push the state of the art.
Further, the performance gap between Table~\ref{tab:compare_1} and~\ref{tab:multitask} indicates the challenge of face recognition under various expressions, which is less studied than pose and illumination variations.

\subsection{How our m-CNN works?}

\begin{figure}[t]
\centering
\begin{tabular}{@{}cc@{}}
\includegraphics[trim=0 0 0 0, clip, width=.23\textwidth]{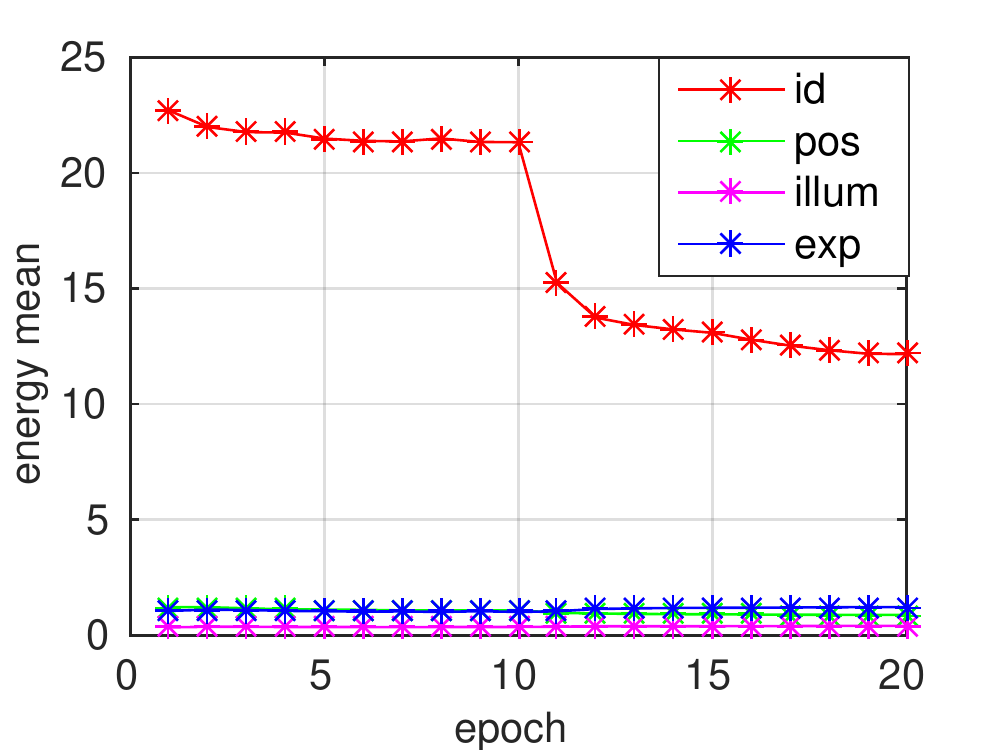} & 
\includegraphics[trim=0 0 0 0, clip, width=.23\textwidth]{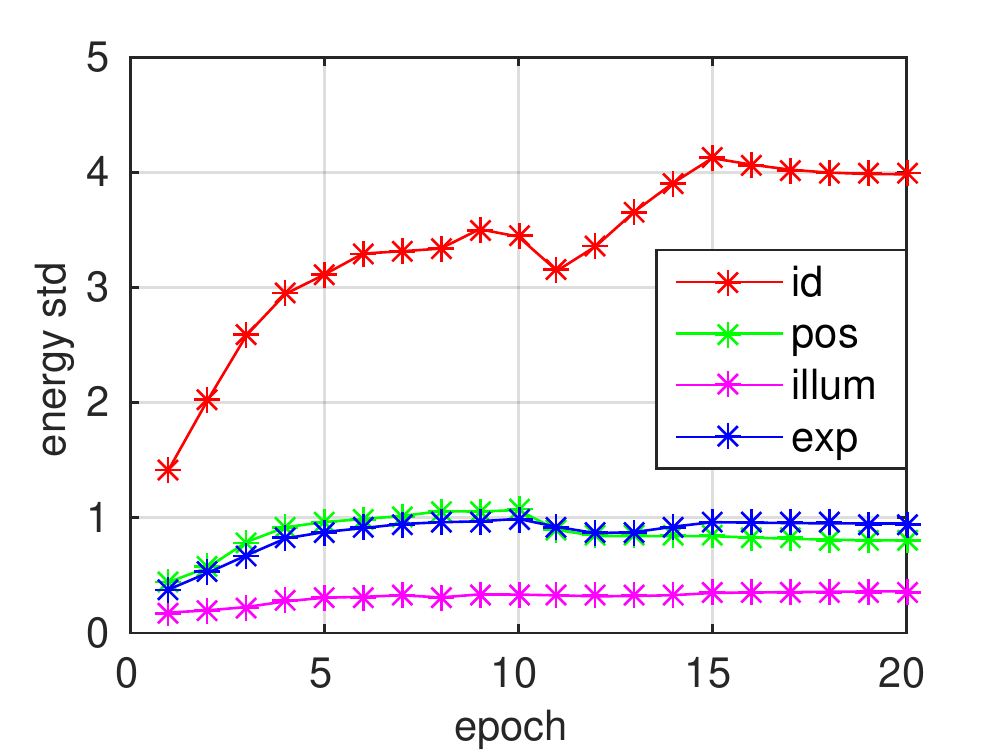} \\ 
\small (a) mean & \small (b) std. \\
\end{tabular}
\caption{The mean and standard deviation of each energy vector during the training process.}
\label{fig:epoch_energy} 
\end{figure}

It is well known in both computer vision and machine learning communities that learning multiple tasks together allows each task to leverage each other and thus improve the generalization ability of the model. 
For CNN-based MTL, previous work~\cite{zhang2016learning} has found that CNN learns shared features for facial landmark localization and the auxiliary tasks including smiling and pose classification. 
This is understandable because the smiling attribute is related to landmark localization as it involves the change of the mouth region. 
However in our case, it is not obvious how the PIE estimations can share features with the main task. 
On contrary, it is more desirable if the learnt identity features are disentangled from the PIE variations. 
Indeed, as will shown later, we have observed that the PIE estimations regularize the CNN to learn PIE-invariant identity features.

We investigate why PIE estimations are helpful for face recognition. 
In our m-CNN (``id+all'' with dynamic weights), all tasks are learnt from the shared representation ${\bf{x}}\in \mathbb{R}^{320\times1}$.
We analyze the importance of each dimension in ${\bf{x}}$ to each task. 
We make the assumption that if a feature dimension is important, the corresponding row in the weight matrix should have high absolute values. 
Taking the main task as an example, we compute an energy vector ${\bf{s}}^d\in \mathbb{R}^{320\times1}$ whose element is computed as:
\begin{equation} 
{\bf{s}}^d_i = \sum_{j=1}^{200} \mid {\bf{W}}^d_{ij}\mid.
\end{equation}

\begin{figure}[t]
\centering
\begin{tabular}{cc}
\includegraphics[width=.23\textwidth]{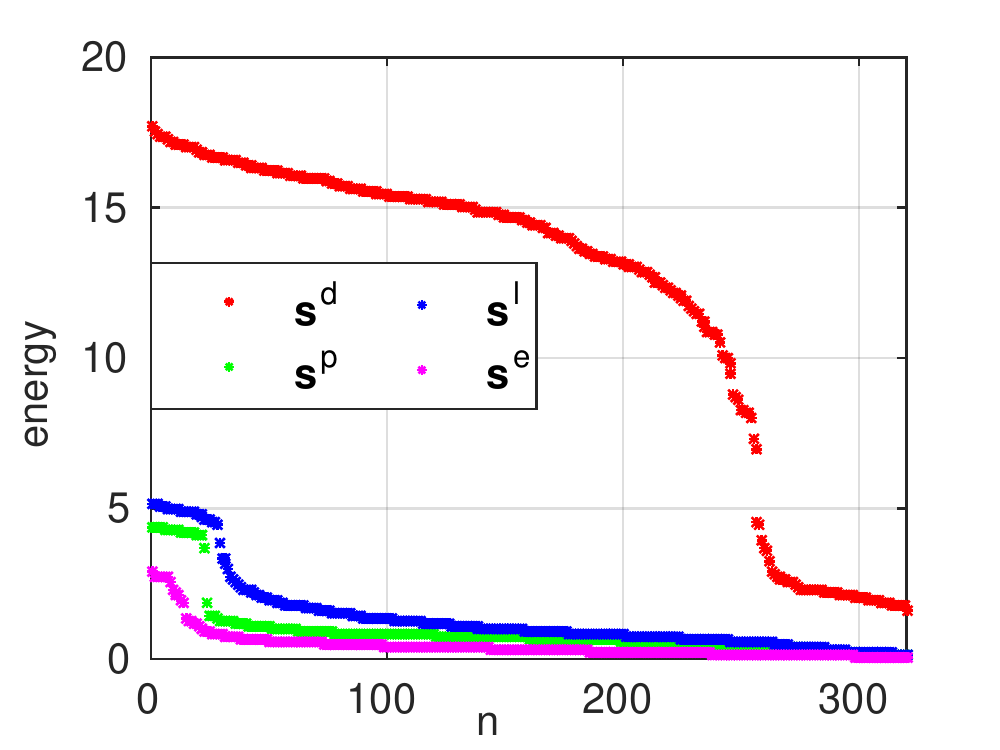} &
\includegraphics[width=.23\textwidth]{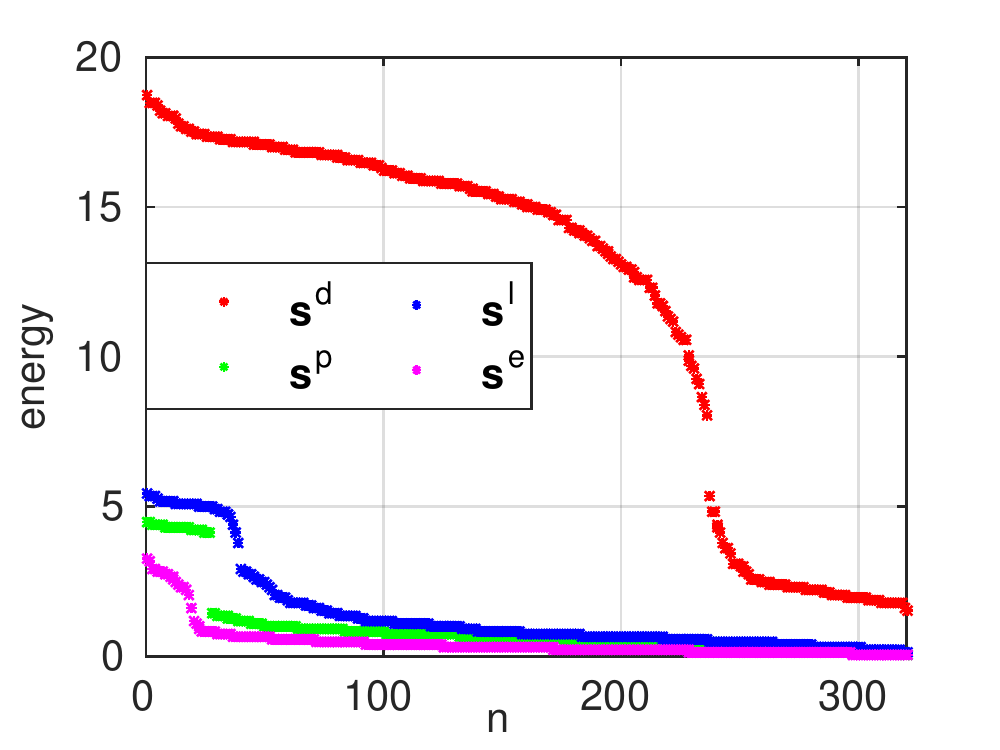} \\
\small (a) $\varphi_s=0.2$ & \small (b) $\varphi_s=0.3$ \\
\end{tabular}
\caption{Energy vectors of m-CNN models with different overall loss weights.}
\label{fig:validation}
\end{figure}

A higher value in ${\bf{s}}^d$ indicates a dimension is more important to face recognition. 
Similarly, we compute the energy vectors for all side tasks as ${\bf{s}}^p$, ${\bf{s}}^l$, ${\bf{s}}^e$ from their weight matrices ${\bf{W}}^p_{320\times 13}$, ${\bf{W}}^l_{320\times 19}$, ${\bf{W}}^e_{320\times 6}$.
We sort each energy vector in a descending order and plot them as shown in Figure~\ref{fig:mCNN} (a). 
The magnitude of the energy vector is linearly correlated to the number of classes in each task. 
The large variance of each energy vector indicates that for each task, each dimension in ${\bf{x}}$ contributes differently. 
Note that the index of the feature dimension is not consistent among them since each energy vector is sorted independently. 

To compare how each dimension contributes to different tasks, we concatenate the weight matrix of all tasks as ${\bf{W}}^{all}_{320\times 238} = [{\bf{W}}^d, {\bf{W}}^p, {\bf{W}}^l, {\bf{W}}^e]$ and compute its energy vector as ${\bf{s}}^{all}$. 
We sort the rows in ${\bf{W}}^{all}$ based on the descending order in energy, as visualized in Figure~\ref{fig:mCNN} (b).
The first $200$ columns represent the sorted ${\bf{W}}^d$ where most energy is distributed in the first $\sim280$ feature dimensions (rows), which are more crucial for face recognition. 
We observe that the shared representation are learnt to allocate a separate set of dimensions for each task, as shown in the block-wise effect in the zoom-in view. 
Each block shows the most essential dimensions with the high energy for PIE estimations respectively. 
Therefore, the shared representation ${\bf{x}}$ can disentangle the PIE variations for identity classification. 

\begin{figure*}[t!]
\centering
\begin{tabular}{cccc}
\includegraphics[width=.23\textwidth]{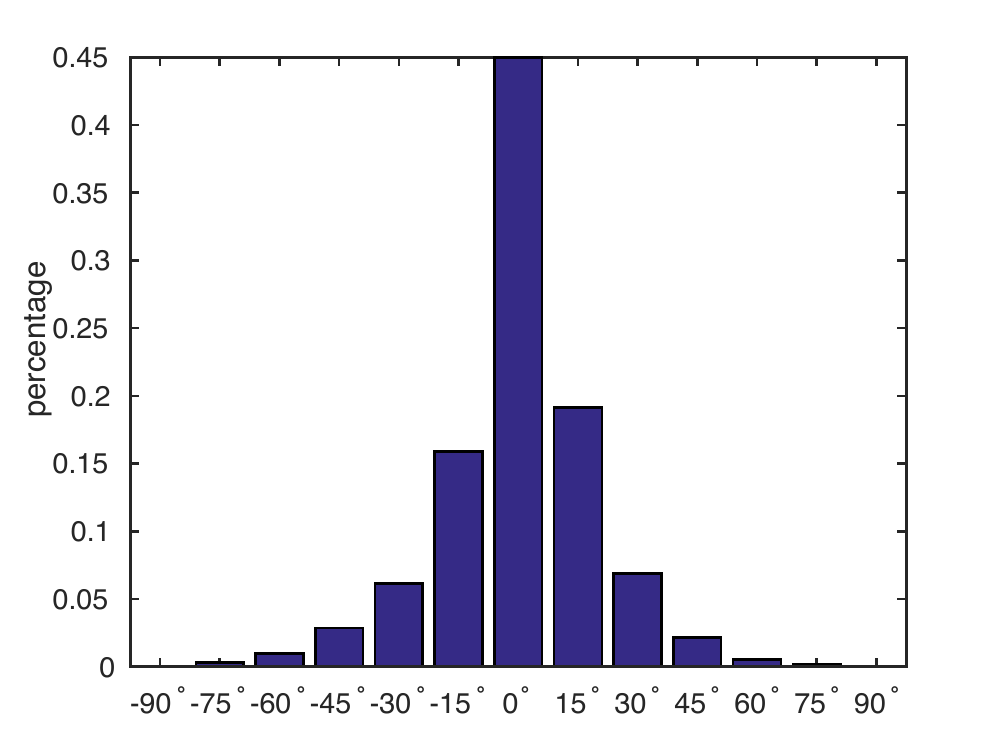} &
\includegraphics[width=.17\textwidth]{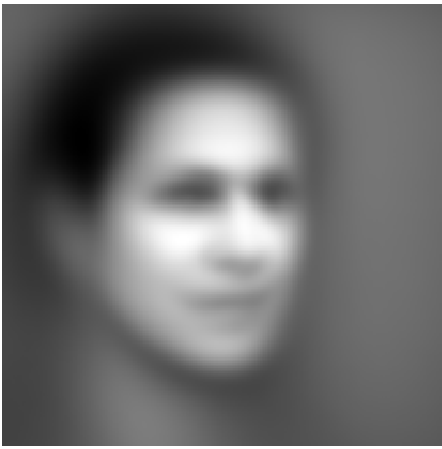} &
\includegraphics[width=.17\textwidth]{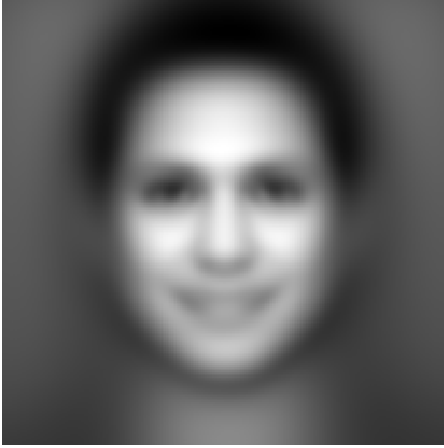} &
\includegraphics[width=.17\textwidth]{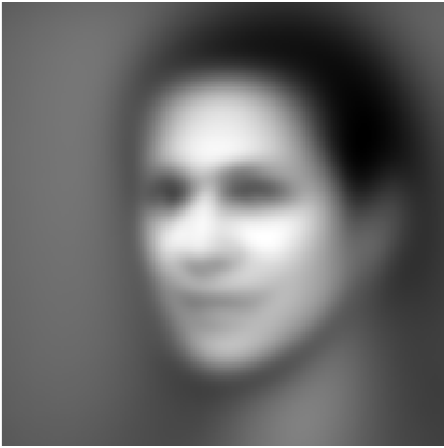} \\
(a) & (b) & (c) & (d) \\
\end{tabular}
\caption{(a) yaw angle distribution on CASIA-Webface; (b) average image of the right profile group; (c) average image of the frontal group; (d) average image of the left profile group.}
\label{fig:meanface}
\end{figure*}

\begin{table*}[t!]
\small
\begin{center}
\caption{Performance comparison on LFW dataset.} 
\label{tab:lfw}
\begin{tabular}{ lclll}
\hline 
Method & $\#$Net & Training Set & Metric & Accuracy $\pm$ Std ($\%$)\\ \hline \hline
DeepID2~\cite{sun2014deep} & $1$ & $202,599$ images of $10,177$ subjects, private & Joint-Bayes & $95.43$\\
DeepFace~\cite{taigman2014deepface} & $1$ & $4.4$M images of $4,030$ subjects, private & cosine & $95.92\pm0.29$\\ 
CASIANet~\cite{yi2014learning} & $1$ & $494,414$ images of $10,575$ subjects, public & cosine & $96.13\pm0.30$ \\
Wang et al.~\cite{wang2016face} & $1$ & $404,992$ images of $10,553$ subjects, public & Joint-Bayes & $96.2\pm0.9$ \\ 
Littwin and Wolf~\cite{littwin2016multiverse} & $1$ & $404,992$ images of $10,553$ subjects, public & Joint-Bayes & $98.14\pm0.19$ \\ 
MultiBatch~\cite{tadmor2016learning} & $1$ & $2.6$M images of $12$K subjects, private & Euclidean & $98.20$ \\
VGG-DeepFace~\cite{parkhi2015deep} & $1$ & $2.6$M images of $2,622$ subjects, public & Euclidean & $98.95$ \\
Wen et al.~\cite{wen2016discriminative} & $1$ & $0.7$M images of $17,189$ subjects, public & cosine & $99.28$ \\ 
FaceNet~\cite{schroff2015facenet} & $1$ & $260$M images of $8$M subjects, private & L$2$ & $99.63\pm0.09$\\ \hline
s-CNN (ours) & $1$ & $494,414$ images of $10,575$ subjects, public & cosine & $97.87\pm0.70$ \\ 
m-CNN (ours) & $1$ & $494,414$ images of $10,575$ subjects, public & cosine & $98.07\pm0.57$ \\ 
p-CNN (ours) & $1$ & $494,414$ images of $10,575$ subjects, public & cosine & $98.27\pm0.64$ \\ \hline\hline
\end{tabular}
\end{center}
\end{table*}

To validate this observation quantitatively, we contrast two types of features for face recognition: 1) a subset of ${\bf{x}}$ with the $n$ largest energy in ${\bf{s}}^d$, which are more crucial in modeling identity variation, 2) the features ${\bf{y}}^d_{200\times1} = {{\bf{W}}^d_{n\times200}}^\intercal {\bf{x}}_{n\times1}+ {\bf{b}}^d$.
We vary $n$ from $100$ to $320$ and compute the rank-$1$ face identification rate on the entire Multi-PIE testing set. 
The performance are shown in Figure~\ref{fig:mCNN} (c). 
When ${\bf{x}}_{n}$ is used, the performance improves with increasing dimensions and drops when additional dimensions are included, which are learnt to model the PIE variations. 
In contrary, the identity features ${\bf{y}}^d$ can eliminate the dimensions that are not helpful for identity classification through the weight matrix ${\bf{W}}^d$, resulting in continuously improved performance w.r.t.~$n$. 
Therefore, the weight matrix ${\bf{W}}^d$ acts like feature selection to select only crucial feature dimensions that are learnt to model identity for PIE-invariant face recognition.

We further analysis how the energy vectors evolve over time during training. 
Specifically, at each epoch, we compute the energy vectors the same way as shown in Figure~\ref{fig:mCNN} (a). 
Then we compute the mean and standard deviation of each energy vector, as shown in Figure~\ref{fig:epoch_energy}. 
Despite some local fluctuations, the overall trend is that the mean is decreasing and standard deviation is increasing as training goes on. 
This is because in the early stage of training, the energy vectors are more evenly distributed among all feature dimensions, which leads to higher mean values and lower standard deviations. 
In the later stage of training, the energy vectors are shaped in a way to focus on some key dimensions for each task and ignore other dimensions, which leads to lower mean values and higher standard deviations. 

CNN learns to allocate a separate set of dimensions in the shared features to each task. 
And how many dimensions are assigned to each task depends on the loss weights. 
Recall that we obtain the overall loss weight for the side tasks as $\varphi_s=0.1$ via brute-force search. 
Figure~\ref{fig:validation} shows the energy distributions with $\varphi_s=0.2$ and $\varphi_s=0.3$, comparing to Figure~\ref{fig:mCNN} (a) where $\varphi_s=0.1$. 
We have two observations. 
First, a larger loss weight for the side tasks leads to more dimensions being assigned to the side tasks. 
Second, the energies in ${\bf{s}}^d$ increase in order to compensate the fact that the dimensions assigned to the main task are decreased. 
Therefore, we conclude that the loss weights control the energy distribution between different tasks. 

\subsection{Unconstrained Face Recognition}
\Paragraph{Experimental Settings}
We use CASIA-Webface~\cite{yi2014learning} as our training set and evaluate on LFW~\cite{huang2007labeled}, CFP~\cite{sengupta2016frontal}, and IJB-A~\cite{klare2015pushing} datasets. 
CASIA-Webface consists of $494,414$ images of $10,575$ subjects. 
LFW consists of $10$ folders each with $300$ same-person pairs and $300$ different-person pairs. 
Given the saturated performance of LFW mainly due to its mostly frontal view faces, CFP and IJB-A are introduced for large-pose face recognition. 
CFP is composed of $500$ subjects with $10$ frontal and $4$ profile images for each subject. 
Similar to LFW, CFP includes $10$ folders, each with $350$ same-person pairs and $350$ different-person pairs, for both frontal-frontal (FF) and frontal-profile (FP) verification protocols. 
IJB-A dataset includes $5,396$ images and $20,412$ video frames of $500$ subjects. 
It defines template-to-template matching for both face verification and identification. 

\begin{table*}[t!]
\small
\begin{center}
\caption{Performance comparison on CFP. Results reported are the average $\pm$ standard deviation over the $10$ folds.} 
\label{tab:cfp}
\begin{tabular}{ lccc|ccc}
\hline 
Method $\downarrow$ & \multicolumn{3}{c}{Frontal-Frontal} & \multicolumn{3}{c}{Frontal-Profile} \\ \hline 
Metric ($\%$) $\to$ & Accuracy & EER & AUC & Accuracy & EER & AUC \\ \hline\hline
Sengupta et al.~\cite{sengupta2016frontal} & $96.40\pm0.69$ & $3.48\pm0.67$ & $99.43\pm0.31$ & $84.91\pm1.82$ & $14.97\pm1.98$ & $93.00\pm1.55$ \\
Sankarana. et al.~\cite{sankaranarayanan2016triplet} & $96.93\pm0.61$& $2.51\pm0.81$ & $99.68\pm0.16$ & $89.17\pm2.35$ & $8.85\pm0.99$ & $97.00\pm0.53$ \\ 
Chen, et al.~\cite{chen2016fisher} & ${\bf{98.67}}\pm0.36$ & ${\bf{1.40}}\pm0.37$ & ${\bf{99.90}}\pm0.09$ & $91.97\pm1.70$ & $8.00\pm1.68$ & $97.70\pm0.82$ \\
DR-GAN~\cite{tran2017disentangled}& $97.84\pm0.79$ & $2.22\pm0.09$ & $99.72\pm0.02$ & $93.41\pm1.17$ & $6.45\pm0.16$ & $97.96\pm0.06$ \\
Human & $96.24\pm0.67$ & $5.34\pm1.79$ & $98.19\pm1.13$ & $94.57\pm1.10$ & $5.02\pm1.07$ & $98.92\pm0.46$ \\ \hline
s-CNN (ours) & $97.34\pm0.99$ & $2.49\pm0.09$ & $99.69\pm0.02$ & $90.96\pm1.31$ & $8.79\pm0.17$ & $96.90\pm0.08$ \\
m-CNN (ours) & $97.77\pm0.39$ & $2.31\pm0.06$ & $99.69\pm0.02$ & $91.39\pm1.28$ & $8.80\pm0.17$ & $97.04\pm0.08$ \\
p-CNN (ours) & $97.79\pm0.40$ & $2.48\pm0.07$ & $99.71\pm0.02$ & ${\bf{94.39}}\pm1.17$ & ${\bf{5.94}}\pm0.11$ & ${\bf{98.36}}\pm0.05$ \\ \hline\hline
\end{tabular}
\end{center}
\end{table*}

\begin{table*}[t]
\small
\begin{center}
\caption{Performance comparison on IJB-A. }
\label{tab:ijb-a}
\begin{tabular}{lcccc}
\hline 
Method $\downarrow$ & \multicolumn{2}{c}{Verification} & \multicolumn{2}{c}{Identification} \\ \hline 
Metric ($\%$) $\to$ & @FAR=$0.01$ & @FAR=$0.001$ & @Rank-$1$ & @Rank-$5$ \\ \hline\hline
OpenBR~\cite{klare2015pushing} & $23.6\pm0.9$ & $10.4\pm1.4$ & $24.6\pm1.1$ & $37.5\pm0.8$ \\
GOTS~\cite{klare2015pushing} & $40.6\pm1.4$ & $19.8\pm0.8$ & $44.3\pm2.1$ & $59.5\pm2.0$ \\
Wang et al.~\cite{wang2016face} & $72.9\pm3.5$ & $51.0\pm6.1$ & $82.2\pm2.3$ & $93.1\pm1.4$ \\ 
PAM~\cite{masi2016pose} & $73.3\pm1.8$ & ${\bf{55.2}}\pm3.2$ & $77.1\pm1.6$ & $88.7\pm0.9$ \\
DR-GAN~\cite{tran2017disentangled} & $77.4\pm2.7$ & $53.9\pm4.3$ & $85.5\pm1.5$ & ${\bf{94.7}}\pm1.1$ \\ 
DCNN~\cite{chen2016unconstrained} & ${\bf{78.7}}\pm4.3$ & \textendash & $85.2\pm1.8$ & $93.7\pm1.0$ \\ \hline
s-CNN (ours) & $75.6\pm3.5$ & $52.0\pm7.0$ & $84.3\pm1.3$ & $93.0\pm0.9$ \\
m-CNN (ours) & $75.6\pm2.8$ & $51.6\pm4.5$ & $84.7\pm1.0$ & $93.4\pm0.7$ \\
p-CNN (ours) & $77.5\pm2.5$ & $53.9\pm4.2$ & ${\bf{85.8}}\pm1.4$ & $93.8\pm0.9$ \\ \hline\hline
\end{tabular}
\end{center}
\end{table*}

In order to apply the proposed m-CNN and p-CNN, we need to have the labels for the side tasks. 
However, it is not easy to manually label our training set. 
Instead, we only consider pose estimation as the side task and use the estimated pose as the label for training. 
We use PIFA~\cite{jourabloo2016large} to estimate $34$ landmarks and the yaw angle, which defines three groups: right profile [$-90^\circ, -30^\circ$), frontal [$-30^\circ, 30^\circ$], and left profile ($30^\circ, 90^\circ$].
Figure~\ref{fig:meanface} shows the distribution of the yaw angle estimation and the average image of each pose group. 
CASIA-Webface is biased towards frontal faces with $88\%$ faces belonging to the frontal pose group based on our pose estimation. 
The sharpness of the average images indicates our pose estimation is accurate to some extent. 

The network structures are similar to those experiments on Multi-PIE. 
All models are trained from scratch for $15$ epochs with a batch size of $8$. 
The initial learning rate is set to $0.01$ and reduced at $10$th and $14$th epoch with a factor of $0.1$.
The other parameter settings and training process are the same as those in Multi-PIE.
We use the same pre-processing as in~\cite{yi2014learning} to align a face image.
Each image is horizontally flipped for data augmentation in the training set. 
We also generate the mirror image of an input face in the testing stage. 
We use the average cosine distance of all four comparisons between the image pair and its mirror images for face comparison. 
Note that we do not use this data augmentation for Multi-PIE because each subject is captured in a full pose range. 
Therefore the mirror image is likely to be similar to another image with a different pose in the dataset. 

\Paragraph{Performance on LFW}
Table~\ref{tab:lfw} compares our face verification performance with state-of-the-art methods on LFW dataset. 
We follow the unrestricted with labeled outside data protocol. 
Although it is well-known that an ensemble of multiple networks can improve the performance~\cite{sun2015deepid3, sun2015deeply}, we only compare CNN-based methods with one network to focus on the power of a single model. 
Our implementation of the CASIA-Net (s-CNN) with BN achieves much better results compared to the original performance~\cite{yi2014learning}. 
Even with such a high baseline, m-CNN and p-CNN can still improve, achieving comparable results with state of the art, or better results if comparing to those methods trained with the same amount of data.
Since LFW is biased towards frontal faces, we expect the improvement of our proposed m-CNN and p-CNN to the baseline s-CNN to be larger if they are tested on large-pose datasets.

\Paragraph{Performance on CFP}
Table~\ref{tab:cfp} shows our face verification performance comparison with state-of-the-art methods on CFP dataset. 
For FF setting, m-CNN and p-CNN improve the verification rate of s-CNN slightly. 
This is expected, as there is little pose variation. 
For FP setting, p-CNN substantially outperforms s-CNN and prior work, reaching close-to-human performance ($94.57\%$). 
Note our accuracy of $94.39\%$ is $14.8\%$ relative error reduction of the previous state of the art DR-GAN with $93.41\%$.
Therefore, the proposed divide-and-conquer scheme is very effective in in-the-wild face verification with large pose variation. 
And the proposed stochastic routing scheme improves the robustness of the algorithm. 
Even with the estimated pose serving as the ground truth pose label for MTL, the models can still disentangle the pose variation from the learnt identity features for pose-invariant face verification. 

\Paragraph{Performance on IJB-A}
We conduct close-set face identification and face verification on IJB-A dataset. 
First, we retrain our models after removing $26$ overlapped subjects between CASIA-Webface and IJB-A. 
Second, we fine-tune the retrained models on the IJB-A training set of each fold for $50$ epochs. 
Similar to~\cite{wang2016face}, we separate all images into ``well-aligned" and ``poorly-aligned" faces based on the face alignment results and the provided annotations. 
In the testing stage, we only select images from the ``well-aligned" faces for recognition. 
If all images in a template are ``poorly-aligned" faces, we select the best aligned face among them. 
Table~\ref{tab:ijb-a} shows the performance comparison on IJB-A. 
Similarly, we only compare to the methods with a single model. 
The proposed p-CNN achieves comparable performance in both face verification and identification. 
The margin between s-CNN and p-CNN shows the merit of MTL for in-the-wild face recognition.

\section{Conclusions}
This paper explores multi-task learning for face recognition with PIE estimations as the side tasks. 
We propose a dynamic-weighting scheme to automatically assign the loss weights for each side task during training. 
MTL helps to learn more discriminative identity features by disentangling the PIE variations. 
We also propose a pose-directed multi-task CNN with stochastic routing scheme to direct different paths for face images with different poses.
We make the first effort to study face identification on the entire Multi-PIE dataset with full PIE variations. 
Extensive experiments on Multi-PIE show that our m-CNN and p-CNN can dramatically improve face recognition performance, especially on large poses. 
The proposed method is applicable to in-the-wild datasets with estimated pose serving as the label for training.
We have achieved state-of-the-art performance on LFW, CFP, and IJB-A, showing the value of MTL for pose-invariant face recognition in the wild.

{\small
\bibliographystyle{ieee}
\bibliography{egbib}
}

\ifCLASSOPTIONcaptionsoff
  \newpage
\fi

\begin{IEEEbiography}[{\includegraphics[width=1in,height=1.25in,clip,keepaspectratio]{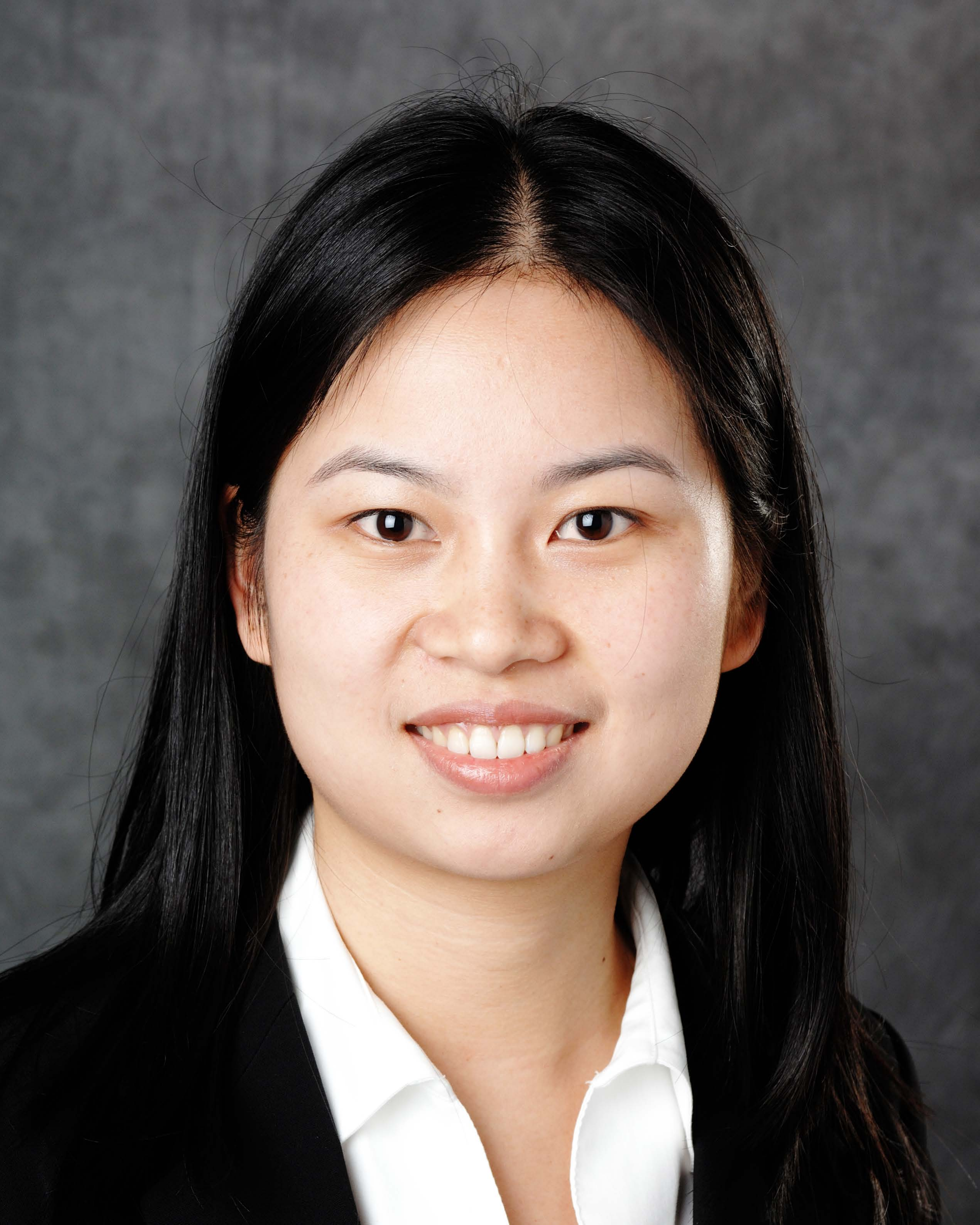}}]
{Xi Yin} received the B.S. degree in Electronic and Information Science from Wuhan University, China, in 2013. Since August 2013, she has been working toward her Ph.D. degree in the Department of Computer Science and Engineering, Michigan State University, USA. Her research area are face recognition, deep learning, and plant image processing. Her paper on multi-leaf segmentation won the Best Student Paper Award at Winter Conference on Application of Computer Vision (WACV) 2014. 
\end{IEEEbiography}

\begin{IEEEbiography}[{\includegraphics[width=1in,height=1.25in,clip,keepaspectratio]{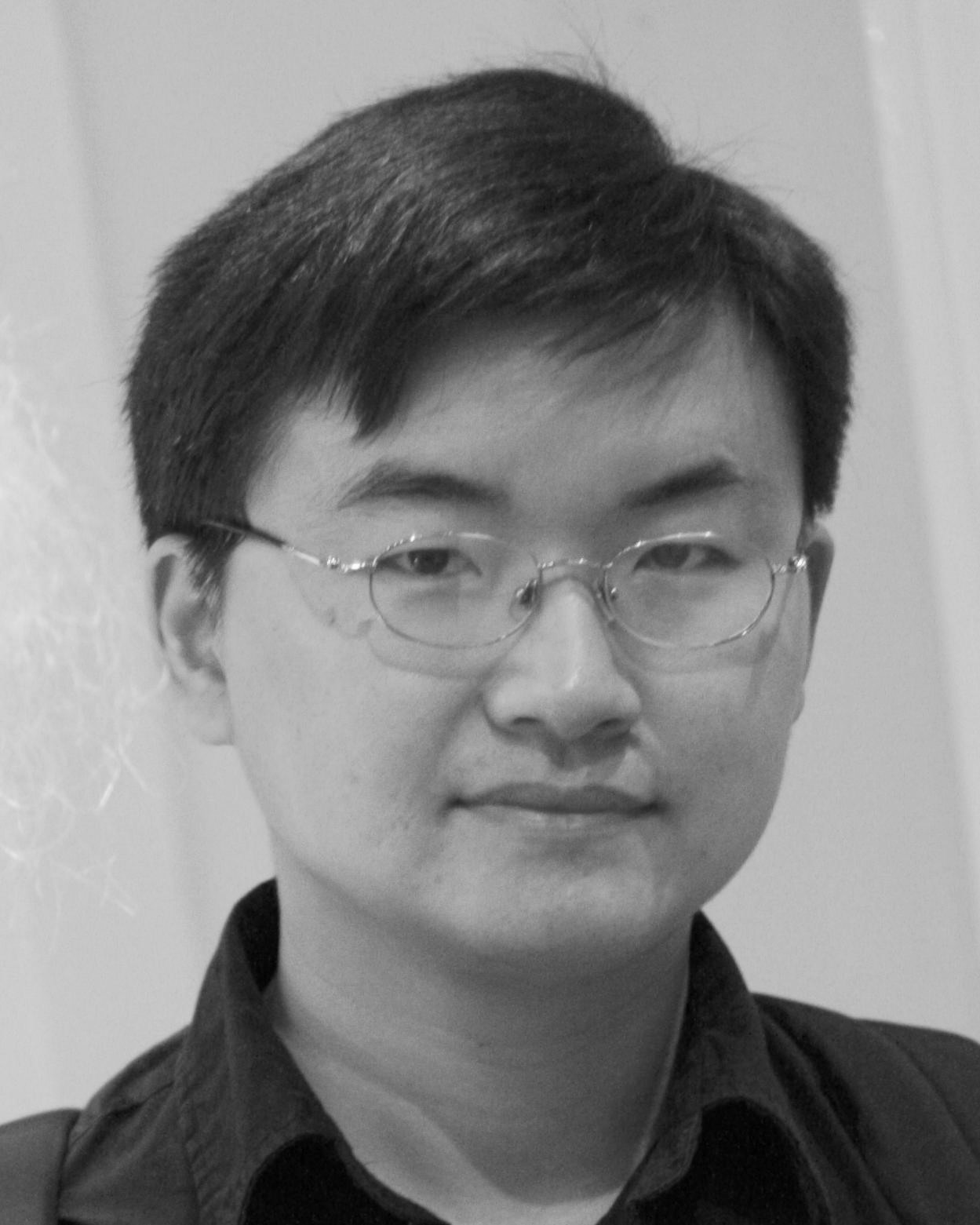}}]
{Xiaoming Liu}is an Assistant Professor at the Department of Computer Science and Engineering of Michigan State University. He received the Ph.D. degree in Electrical and Computer Engineering from Carnegie Mellon University in 2004. Before joining MSU in Fall 2012, he was a research scientist at General Electric (GE) Global Research. His research interests include computer vision, machine learning, and biometrics. As a co-author, he is a recipient of Best Industry Related Paper Award runner-up at ICPR 2014, Best Student Paper Award at WACV 2012 and 2014, and Best Poster Award at BMVC 2015. He has been the Area Chair for numerous conferences, including FG, ICPR, WACV, ICIP, and CVPR. He is the program chair of WACV 2018. He is an Associate Editor of Neurocomputing journal. He has authored more than 100 scientific publications, and has filed 22 U.S. patents. 
\end{IEEEbiography}

\end{document}